\pdfoutput=1
\documentclass[11pt,a4paper]{article}
\usepackage[final]{acl2023}
\usepackage{float}
\usepackage{times}
\usepackage{amsmath}
\usepackage{latexsym}
\usepackage[normalem]{ulem}
\usepackage{graphicx}
\usepackage{bm}
\usepackage{subcaption, booktabs} 
\usepackage{xcolor}
\usepackage{soulutf8}
\usepackage{multirow}
\usepackage{CJKutf8}
\usepackage{arabtex}
\usepackage[T2A,LAE,T1]{fontenc}
\usepackage[russian,USenglish]{babel}
\usepackage{array}
\usepackage{color}
\usepackage{listings}
\definecolor{lightgreen}{RGB}{194, 245, 186}
\definecolor{darkgreen}{RGB}{80, 182, 39}
\definecolor{lightblue}{RGB}{163, 223, 255}
\usepackage{ulem}
\usepackage{utf8}
\usepackage{titlesec}
\setcode{utf8}
\usepackage{caption}
\titlespacing{\paragraph}{%
  0pt}{%              left margin
  0.1\baselineskip}{% space before (vertical)
  0.7em}%               space after (horizontal)

\newcommand\hlpink{\bgroup\markoverwith
  {\textcolor{pink}{\rule[-.5ex]{2pt}{2.5ex}}}\ULon}
\newcommand\hlblue{\bgroup\markoverwith
  {\textcolor{lightblue}{\rule[-.5ex]{2pt}{2.5ex}}}\ULon}
\newcommand\hlgreen{\bgroup\markoverwith
  {\textcolor{lightgreen}{\rule[-.5ex]{2pt}{2.5ex}}}\ULon}

\makeatletter
\gdef\Shortstack{\@ifnextchar[\@Shortstack{\@Shortstack[c]}}
\gdef\@Shortstack[#1]#2{%
  \leavevmode
  \vbox\bgroup
    \baselineskip-\p@\lineskip 3\p@
    \let\mb@l\hss\let\mb@r\hss
    \expandafter\let\csname mb@#1\endcsname\relax
    \let\\\@stackcr\setlength{\baselineskip}{#2}%
    \@ishortstack}
\makeatother
\newcolumntype{P}[1]{>{\centering\arraybackslash}p{#1}}
\newcommand{\hlc}[2][yellow]{{%
    \colorlet{foo}{#1}%
    \sethlcolor{foo}\hl{#2}}%
}

\usepackage{microtype}

\newtoggle{draft}
\toggletrue{draft}
\iftoggle{draft}{
    \newcommand{\ky}[1]{\textcolor{violet}{[KY: #1]}}
    \newcommand{\patrick}[1]{\textcolor{blue}{[PF: #1]}}
    \newcommand{\andre}[1]{\textbf{\textcolor{red}{[AM: #1]}}}
    \newcommand{\gn}[1]{\textcolor{magenta}{[GN: #1]}}
    
}{
    \newcommand{\ky}[1]{}
    \newcommand{\patrick}[1]{}
    \newcommand{\andre}[1]{}
    \newcommand{\gn}[1]{}
}

\newcommand*\samethanks[1][\value{footnote}]{\footnotemark[#1]}

\title{When Does Translation Require Context? \\ A Data-driven, Multilingual Exploration}
 
\author{
Patrick Fernandes$^{1,2,3}$\thanks{\,\, Equal contribution} \qquad
Kayo Yin$^{4}$\samethanks \qquad Emmy Liu$^{1}$ \\
\textbf{André F. T. Martins}$^{2,3,5}$\qquad 
\textbf{Graham Neubig}$^{1}$ \\
$^1$Language Technologies Institute, Carnegie Mellon University, Pittsburgh, PA \\
$^2$Instituto Superior Técnico \& LUMLIS (Lisbon ELLIS Unit), Lisbon, Portugal \\
$^3$Instituto de Telecomunicações, Lisbon, Portugal \\
$^4$University of California, Berkeley \qquad $^5$Unbabel, Lisbon, Portugal \\
 {\small \texttt{pfernand@cs.cmu.edu} \quad \texttt{kayoyin@berkeley.edu}}
}
\date{}

\begin{document}
\maketitle
\begin{abstract}
Although proper handling of discourse significantly contributes to the quality of machine translation (MT), these improvements are not adequately measured in common translation quality metrics. Recent works in context-aware MT attempt to target a small set of discourse phenomena during evaluation, however not in a fully systematic way. In this paper, we develop the \textbf{Mu}ltilingual \textbf{D}iscourse-\textbf{A}ware ({\sc MuDA}) benchmark, a series of taggers that identify and evaluate model performance on discourse phenomena in any given dataset. The choice of phenomena is inspired by a novel methodology to systematically identify translations requiring context. We confirm the difficulty of previously studied phenomena while uncovering others that were previously unaddressed.
%In this paper, we propose a methodology to identify translations that require context systematically, and use this methodology to both confirm the difficulty of previously studied phenomena as well as uncover new ones that have not been addressed in previous work. We then develop the \textbf{Mu}ltilingual \textbf{D}iscourse-\textbf{A}ware (MuDA) benchmark, a series of taggers for these phenomena in 14 different language pairs, which we use to evaluate context-aware MT. 
We find that common context-aware MT models make only marginal improvements over context-agnostic models, which suggests these models do not handle these ambiguities effectively. We release code and data for 14 language pairs to encourage the MT community to focus on accurately capturing discourse phenomena.\footnote{Code available at 
\url{https://github.com/CoderPat/MuDA}
%\url{https://anonymous.4open.science/status/MuDA-FCDE}
. See \S\ref{appendix:muda-toolkit} for example usages of our released toolkit} 
%\footnote{\andre{this needs to be anonymized!!!}\href{https://github.com/neulab/contextual-mt}{https://github.com/neulab/contextual-mt}}
\end{list} 
% SUPER HACK with the end list to fix error in abstract

\end{abstract}

\section{Introduction}

%\gn{The title is perhaps a bit too abstract. Maybe ``When Does Translation Require Context? A Data-driven, Multilingual Exploration'' or something like that would be more concrete and focus on the two main contributions.}

%\andre{I saw many occurrences of the work ``phenomena'' in the abstract and intro, should we try to reduce a few? we could use some pronoun phenomena for that ;)}
%\gn{Re-reading the abstract, I agree.}

In order to properly translate discourse phenomena including anaphoric pronouns, lexical cohesion, and discourse markers, a machine translation (MT) model must use information from previous utterances \cite{guillou-etal-2018-pronoun, laubli-etal-2018-machine, toral-etal-2018-attaining}.
%In machine translation (MT), information from previous utterances has been found crucial to adequately translate a number of discourse phenomena including 
%\andre{maybe replace ``such as'' by ``including'', it seemed a bit ackward when I read this to see pronouns an instance of a ``phenomenon''}
%anaphoric pronouns, lexical cohesion, and discourse markers \cite{guillou-etal-2018-pronoun, laubli-etal-2018-machine, toral-etal-2018-attaining}. 

However, while generating proper translations of these phenomena is important for comprehension,
they represent a small portion of words in natural language.
Therefore, common metrics such as BLEU \cite{papineni-etal-2002-bleu} cannot be used to judge the quality of discourse translation.
%do not provide a clear picture of whether they are appropriately captured or not. 

% \begin{figure}[t]
%     \includegraphics[width=\linewidth]{Drawing.png}
%     \caption{\vspace{-1mm}\ky{Ideas for our first figure?} \gn{I think the table serves as a sufficiently good illustration of the main contributions, we probably don't need another figure.}}
%     \vspace{-2mm}
%     \label{fig:main}
% \end{figure}

\begin{table}[t]
\renewcommand{\arraystretch}{0.8}
\resizebox{\linewidth}{!}{ 
\begin{tabular}{c|c|c}
\toprule
Dataset & Lang. & Phenomena \\ 
\midrule
\citet{muller-etal-2018-large} & $\text{EN}\rightarrow \text{DE}$ & Pronouns \\
\midrule
\multirow{2}{*}{\citet{bawden-etal-2018-evaluating}} & \multirow{2}{*}{$\text{EN}\rightarrow \text{FR}$} & Pronouns, Coherence \\ 
& & Lexical Consistency \\
% & & Cohesion \\
% & & Coherence \\
\midrule
\multirow{3}{*}{\Shortstack[c]{1.2em}{\citet{voita-etal-2018-context} \\\citet{voita-etal-2019-good} }} & \multirow{3}{*}{$\text{EN}\rightarrow \text{RU}$} & Pronouns \\
& & Deixis, Ellipsis \\
& & Lexical Consistency \\
\midrule
\multirow{3}{*}{\citet{DBLP:journals/corr/abs-2004-14607}} & \multirow{3}{*}{\shortstack[c]{$\text{DE}\rightarrow \text{EN}$ \\ $\text{FR}\rightarrow \text{EN}$ \\ $\text{RU}\rightarrow \text{EN}$}} & Pronouns, Coherence \\
& & Lexical Consistency \\
& & Discourse Connectives \\
\midrule
\multirow{4}{*}{Our Work} & \multirow{4}{*}{14 Pairs (\S\ref{section:benchmark})} & Pronouns, Ellipsis \\
& & Formality \\ 
& & Lexical Consistency \\
&& Verb Forms \\
\bottomrule
\end{tabular}}
\vspace{-1.5mm}
\caption{Some representative works on contextual machine translation that perform evaluation on discourse phenomena, contrasted to our work. For a more complete review see \citet{10.1145/3441691}.}
\vspace{-6.5mm}
\label{table:phenomena-dataset}
\end{table}

Recent work on neural machine translation
%\andre{does this need to be capitalized?}
 (NMT) models that attempt to incorporate extra-sentential context \citep[\textit{inter alia}]{tiedemann-scherrer-2017-neural, miculicich-etal-2018-document, maruf-haffari-2018-document} often perform targeted evaluation of certain discourse phenomena, mostly focusing on ellipsis, formality \cite{voita-etal-2019-good, voita-etal-2019-context}, and pronoun translation \cite{muller-etal-2018-large, bawden-etal-2018-evaluating, lopes-etal-2020-document}. However,  only a limited set of discourse phenomena for a few language pairs have been studied (see summary in Table \ref{table:phenomena-dataset}).
The difficulty of broadening these studies stems from the reliance of previous work on introspection and domain knowledge to identify the relevant discourse phenomena, frequently involving expert speakers, which then requires engineering complex language-specific methods to create test suites or manually designing data for evaluation.

In this paper, we identify sentences that contain discourse phenomena through a \emph{data-driven, semi-automatic methodology}. We apply this method to create a \emph{multilingual benchmark testing discourse phenomena} in the domain of MT. 
First, we develop P-CXMI (\S\ref{section:cxmi}) as a metric to identify when context is helpful in MT, or more broadly text generation in general.
Then, we perform a systematic analysis of words with high P-CXMI to find categories of translations where context is useful (\S\ref{section:thematic}). We identify novel discourse phenomena that to our knowledge have not been addressed previously (e.g. consistency of verb forms), without requiring \textit{a-priori} language-specific knowledge.
Finally, we design a series of methods to automatically tag words belonging to the identified classes of ambiguities (\S\ref{section:automatic})  and we evaluate existing translation models for different categories of ambiguous translations (\S\ref{section:benchmark}).

We examine a parallel corpus spanning 14 language pairs, measuring translation ambiguity and model performance. We find that the context-aware methods, while improving on standard evaluation metrics, only perform significantly better than context-agnostic baselines for certain discourse phenomena in our benchmark. Our benchmark provides a more fine-grained evaluation of translation models and reveals weaknesses of context-aware models, such as verb form cohesion. We also find that DeepL, a commercial document-level translation system, does better in our benchmark than its sentence-level ablation and Google Translate.
We hope that the released benchmark and code, as well as our findings, will spur targeted evaluation of discourse phenomena in MT to cover more languages and more phenomena in the future.

\begin{table*}[t]
\centering
\vspace{-5mm}
\resizebox{0.95\linewidth}{!}{ 
\begin{tabular}{l|c}
\toprule
\textit{Avelile's mother had HIV virus.} \hlblue{Avelile} had the virus, she was born with the virus. & \multirow{2}{*}{Lexical Cohesion}  \\
\begin{CJK*}{UTF8}{gbsn} \textit{\hlgreen{阿维利尔}的母亲 是携有艾滋病病毒。}
\hlpink{阿维利尔}也有艾滋病病毒。 她一生下来就有。
\end{CJK*} & \\
\midrule
\textit{Your daughter?} \hlc[lightblue]{Your} niece? & Formality \\
 \textit{\hlc[lightgreen]{Votre} fille ?} \hlc[pink]{Votre} nièce ? & (T-V)\\
\midrule
 \textit{Roger. I got'em.} Two-Six, this is Two-Six , we're \hlc[lightblue]{mobile}. & Formality \\
\begin{CJK}{UTF8}{min} \textit{了解 捕捉\hlgreen{した}。} 2-6 こちら移動中\hlpink{だ}。\end{CJK}& (Honorifics)\\
\midrule
 \textit{Our tools today don't look like shovels and picks.} \hlc[lightblue]{They} look like the stuff we walk around with. & \multirow{2}{*}{Pronouns} \\
\textit{As ferramentas de hoje não se parecem com \hlc[lightgreen]{pás e picaretas}.} \hlc[pink]{Elas} se parecem com as coisas que usamos. & \\
\midrule
 \textit{Louis XIV had a lot of people working for him.} They \hlc[lightblue]{made} his silly outfits, like this. & \multirow{2}{*}{Verb Form}  \\
 \textit{Luis XIV \hlc[lightgreen]{tenía} un montón de gente trabajando para él.} Ellos \hlc[pink]{hacían} sus trajes tontos, como éste. & \\
\midrule
 \textit{They're the ones who know what society is going to be like in another generation.} I \hlc[lightblue]{don't}. & \multirow{2}{*}{Ellipsis} \\
\textit{Ancak onlar başka bir nesilde toplumun nasıl olacağını \hlc[lightgreen]{biliyorlar}.} Ben \hlc[pink]{bilmiyorum}. & \\
\bottomrule
\end{tabular}}
\vspace{-1mm}
\caption{Examples of high P-CXMI tokens and corresponding linguistic phenomena. Contextual sentences are \textit{italicized}. The high P-CXMI target token is highlighted in \hlc[pink]{pink}, source and contextual target tokens related to the high P-CXMI token are highlighted in \hlc[lightblue]{blue} and \hlc[lightgreen]{green} respectively.}
\vspace{-2mm}
\label{table:high-cxmi}
\end{table*}

% \begin{table*}[t]
% \resizebox{\linewidth}{!}{ 
% \begin{tabular}{c|cccccccccccccc}
% \toprule
% & ar & de &	es&	fr&	he&	it&	ja&	ko&	nl&	pt&	ro&	ru&	tr&	zh \\
% \midrule
% no tag & 75.72 & 85.52 & 86.24 & 83.55 & 81.09 & 84.32 & 69.83 & 60.85 & 82.86 & 84.17 & 82.55 & 75.9 & 74.27 & 65.82 \\
% with tag & 24.28 & 14.48 & 13.76 & 16.45 & 18.91 & 15.68 & 30.17 & 39.15 & 17.14 & 15.83 & 17.45 & 24.1 & 25.73 & 34.18 \\
% \midrule
% ellipsis & 6.49 & 2.09 & 1.62 & 2.74 & 6.2 & 2.36 & 10.21 & 10.81 & 2.29 & 2.02 & 4.2 & 4.09 & 10.71 & 7.36 \\
% formality & 0.0 & 1.49 & 0.58 & 1.13 & 0.0 & 0.59 & 2.02 & 3.43 & 1.24 & 0.84 & 0.64 & 0.81 & 0.11 & 0.84 \\
% lexical & 1.57 & 2.04 & 2.28 & 2.13 & 1.88 & 2.03 & 1.38 & 0.76 & 2.51 & 2.68 & 1.53 & 1.17 & 1.38 & 1.7 \\
% polysemous & 13.78 & 0.48 & 9.56 & 9.08 & 11.33 & 11.01 & 12.55 & 22.64 & 11.12 & 10.19 & 11.04 & 14.92 & 13.98 & 15.67 \\
% pronouns & 0.27 & 0.83 & 0.1 & 0.55 & 0.0 & 0.0 & 0.38 & 0.0 & 0.0 & 0.53 & 0.16 & 0.0 & 0.0 & 0.0 \\
% verb form & 7.28 & 9.16 & 0.67 & 2.55 & 0.68 & 0.78 & 10.28 & 15.99 & 1.64 & 0.79 & 1.24 & 8.69 & 0.14 & 17.39 \\
% \bottomrule
% \end{tabular}}
% \vspace{-1mm}
% \caption{Percentage of tokens that have been tagged} 
% \vspace{-2mm}
% \label{table:tag_percent}
% \end{table*}

\section{Measuring Context Usage}
\label{section:cxmi}
\subsection{Cross-Mutual Information}

Past work on contrastive evaluation has examined correct and incorrect translations of specific discourse phenomena \cite{bawden-etal-2018-evaluating, muller-etal-2018-large}, but this provides only a limited measure of context usage on phenomena defined by the creators of the dataset. We are therefore interested in devising a metric that is able to capture \textit{all} context usage by a model, beyond a predefined set. 

Conditional Cross-Mutual Information (CXMI) \cite{bugliarello-etal-2020-easier, fernandes21acl} measures the influence of context on model predictions at the corpus level. CXMI is defined as:
\begin{align*}
\text{\textsc{CXMI}}(C\rightarrow Y|X) &=   \\
& \hspace{-2em} \text{H}_{q_{MT_A}}(Y|X) - \text{H}_{q_{MT_C}}(Y|X,C),
\end{align*}
where $X$ and $Y$ are a source and target sentence, respectively, $C$ is the context, $\text{H}_{q_{MT_A}}$ is the entropy of a \textit{context-agnostic} MT model,  and $\text{H}_{q_{MT_C}}$ refers to a  \textit{context-aware} MT model. This quantity can be estimated over a held-out set with $N$ sentence pairs and their respective context as:
\begin{align*}
\text{\textsc{CXMI}}(C\rightarrow Y|X) & \approx \\
& \hspace{-2em} -\frac{1}{N} \sum_{i=1}^N \log \frac{q_{MT_A}(y^{(i)}|x^{(i)})}{q_{MT_C}(y^{(i)}|x^{(i)},C^{(i)})}
\end{align*}

Importantly, the authors find that training a \textit{single} model $q_{MT}$ as both the context-agnostic and context-aware model ensures that non-zero CXMI values are due to context and not other factors (see \citet{fernandes21acl} and \S\ref{sec:data-model-thematic} for details).

Although this approach is promising, it is defined only at a \textit{corpus level}: as the previous equation shows, CXMI is estimated by over a full set of sentences. Since we are interested in measuring how important context is for single sentences or words within a sentence, we extend this definition to capture lower-level context dependency in the next section.

\subsection{Context Usage Per Sentence and Word}

%CXMI measures the context usage by a model by comparing the log-likelihood ratio of samples across \textit{the whole corpus}.
%However, for our purposes, we are interested in measuring how much the context is helpful for single sentences or even just particular words in a sentence.
Pointwise Mutual Information (P-MI) \cite{church-hanks-1990-word} measures the association between two random variables for \textit{specific} outcomes. Mutual information can be seen as the expected value of P-MI over all possible outcomes of the variables. 

Taking inspiration from this, we define the \textbf{Pointwise Cross-Mutual Information} (P-CXMI) for a source, target, context triplet $(x, y, C)$ as:
\begin{align*}
\text{\textsc{P-CXMI}}(y, x, C) = - \log \frac{q_{MT_A}(y|x)}{q_{MT_C}(y|x,C)}
\end{align*}

Intuitively, P-CXMI measures how much more (or less) likely a target sentence $y$ is when it is given context $C$, compared to not being given that context. Note that this is estimated \textit{according to the models $q_{MT_A}$ and $q_{MT_C}$} since, just like CXMI, this measure depends on their learned distributions.

We can also apply P-CXMI at the \textit{word level} to measure how much more likely a particular word in a sentence is when it is given the context, by leveraging the auto-regressive property of the neural decoder. Given the triplet $(x, y, C)$ and the word index $i$, we can measure the P-CXMI for that particular word as:
\begin{align*}
\text{\textsc{P-CXMI}}(i, y, x, C) = -\log \frac{q_{MT_A}(y_i|y_{t<i}, x)}{q_{MT_C}(y_i|y_{t<i},x,C)}
\end{align*}
Note that nothing constrains the form of $C$ or even $x$ and P-CXMI can, in principle, be applied to any conditional language modelling problem.

We use this metric to find words that are strongly context-dependent, which is to say that their likelihood increases greatly with context relative to other words. These words are the ones that likely correspond to discourse phenomena.

%Using this metric, we now ask: what kind of words tend to see their likelihood increase when given the context?
%Such words should have a high P-CXMI, which we examine in the following \S\ref{section:thematic}.

\section{Which Translation Phenomena Benefit from Context?}
\label{section:thematic}

%\el{I'm not sure about the difference between vocabulary items and individual tokens here. Is vocabulary items just full words?}\ky{Yeah I agree this should be made clearer here! Vocabulary item = we analyze words in aggregate over the corpus (so we essentially look at mean P-CXMI over all occurrences of each word), and individual token = we look at a specific word in context of the sentence it appears in (we look at the P-CXMI of one word) }
To identify salient translation phenomena that require context, we perform a \emph{thematic analysis} \cite{braun2006using}, examining words with high P-CXMI across different language pairs and manually identifying patterns and categorizing them into phenomena where context is useful for translation.

To do so, we systematically examined (1) the mean P-CXMI per part-of-speech (POS) tag, (2) the words with the highest mean P-CXMI across the corpus, and (3) the individual words with the highest P-CXMI in a particular sentence.

\subsection{Data \& Model}
\label{sec:data-model-thematic}

%To compare linguistic phenomena that arise during document-level translation across various language pairs, we need a dataset that is document-level, rich in context-dependent discourse phenomena, and parallel in multiple languages. We, therefore, perform our study on transcripts of TED talks and their translations \cite{qi-etal-2018-pre}. We choose to study translation between English and Arabic, German, Spanish, French, Hebrew, Italian, Japanese, Korean, Dutch, Portuguese, Romanian, Russian, Turkish and Mandarin Chinese. These 14 target languages are chosen for their high availability of TED talks and linguistic tools, as well as for the diversity of language types in our comparative study (Table \ref{table:langs} in Appendix~\ref{appendix:lang}). For each language pair, our dataset contains 113,711 parallel training sentences from 1,368 talks, 2,678 development sentences from 41 talks, and 3,385 testing sentences from 43 talks.

To compare linguistic phenomena that arise during document-level translation across language pairs, we use a dataset consisting of TED talks' transcripts and translations \cite{qi-etal-2018-pre}. We use this dataset due to its abundance of discourse phenomena, as well as its availability across many parallel languages. We study translation between English and Arabic, German, Spanish, French, Hebrew, Italian, Japanese, Korean, Dutch, Portuguese, Romanian, Russian, Turkish and Mandarin Chinese. These 14 target languages are chosen for their high availability of TED talks and linguistic tools, as well as for the diversity of language types in our comparative study (Table \ref{table:langs} in Appendix~\ref{appendix:lang}). For each language pair, our dataset contains 113,711 parallel training sentences from 1,368 talks, 2,678 development sentences from 41 talks, and 3,385 testing sentences from 43 talks.

%\el{I find the dataset definition unclear, is it 113711 sentences for each pair exactly? Guessing that would be due to having the same set of talks with parallel data in many languages for each talk?}
%\ky{Yep, it's 113711 sentences in each of the 14 languages and they're parallel across all 14 languages - does this need to be clarified in the writing?}
To obtain the P-CXMI for words in the data, we train a small Transformer \cite{46201} model for every target language and incorporate the target context by concatenating it with the current target sentence  \cite{tiedemann-scherrer-2017-neural}. We train the model with \textit{dynamic} context size \citep{fernandes21acl}, by sampling 0-3 target context sentences and estimating P-CXMI by using this model for $q_{MT_A}$ and $q_{MT_C}$ (details in Appendix~\ref{appendix:training_details}).

\subsection{Analysis Procedure}

We start our analysis by studying POS tags with high mean P-CXMI. In Appendix \ref{appendix:all_cxmi}, we report the mean P-CXMI for selected POS tags on test data. Some types of ambiguity, such as dual form pronouns (\S\ref{section:analysis-results}), can be linked to a single POS tag and be identified at this step, whereas others require finer inspection. 

Next, we inspect the vocabulary items with high mean P-CXMI. At this step, we can detect phenomena that are reflected by certain lexical items that consistently benefit from context for translation.

Finally, we examine individual tokens that obtain the highest P-CXMI. In doing so, we identify patterns that do not depend on lexical features, but rather on syntactic constructions for example. In Table \ref{table:high-cxmi}, we provide selected examples of tokens that have high P-CXMI and the discourse phenomenon we have identified from them. 

\subsection{Identified Phenomena}
\label{section:analysis-results}
%\el{I reorganized this a bit to make definitions of the phenomena more clear}

Through our thematic analysis of items with high P-CXMI, we identified various types of translation ambiguity. Unlike previous work, our method requires no prior knowledge of languages and easily scales to new languages (\S\ref{sec:extension}). 

Although this procedure may find phenomena that are intuitive to the annotators, the data-driven approach makes confirmation bias less severe than works relying on introspection. Hence, our procedure can allow us to discover relevant phenomena that have not been previously addressed, such as verb forms. Examples of each phenomenon are given in \autoref{table:high-cxmi}.

% First, we find relatively high P-CXMI for proper nouns (PROPN) for most languages. As in the first row of Table \ref{table:high-cxmi}, proper nouns may have multiple possible translations
% %(another translation for ``Avelile'' is\begin{CJK*}{UTF8}{gbsn}``艾薇儿''\end{CJK*})
% , but the same entity should be referred to by the same word in a translated document for \textbf{lexical cohesion} 
% %\andre{in the previous section we were using mostly italics to emphasize important words (such as P-CXMI) and here we use bold. Unify? (I like bold, if we don't use too much of it).}
% \cite{carpuat-2009-one}. 
\subsubsection{Lexical Cohesion}
Entities may have multiple possible translations in the target language, but the same entity should be referred to by the same word in a translated document. This is called lexical cohesion.

%In the example, "Avelile" has another possible translation, \begin{CJK*}{UTF8}{gbsn}``艾薇儿''\end{CJK*}, but in order to match the translation in the previous sentence, the translation should be \begin{CJK*}{UTF8}{gbsn}``阿维利尔''\end{CJK*}.

\subsubsection{Formality}
We identify two phenomena which fall under the general category of formality. First, several languages we examined have a T-V distinction (Appendix \ref{appendix:lang}, ``Pronouns Politeness'') in which the second-person pronouns a speaker uses to refer to someone depend on the relationship between the speaker and the addressee. 
%In the example, "votre" is the more formal second-person pronoun.

Second, languages such as Japanese and Korean use honorifics to indicate formality, which are special titles or words expressing courtesy or respect for position. 

\subsubsection{Pronoun Choice}
Unlike in English, many languages use gendered pronouns for pronouns other than the third-person singular, or assign gender based on formal rules rather than semantic ones. In order to assign the correct pronoun, it is therefore necessary to use the previous context to distinguish the grammatical gender of the antecedent.

\subsubsection{Verb Form}
While English verbs may have five forms \footnote{(e.g. \textit{write, writes, wrote, written, writing})}, other languages may have a more fine-grained verb morphology. For example, English has only a single form for the past tense, while the Spanish past tense consists of six verb forms. Verbs must be translated using the verb form that reflects the tone, mood and cohesion of the document.

\subsubsection{Ellipsis}
Ellipsis refers to the omission of superfluous words that are able to be inferred from the context. For instance, in the last row of \autoref{table:high-cxmi}, 
the English text does not repeat the verb \textit{know} in the second sentence as it can be understood from the previous sentence. However, in Turkish, there is no natural way to translate the verb-phrase ellipsis, so context is important for translating the verb correctly.

\section{Cross-phenomenon MT Evaluation}
\label{section:automatic}

Next, we we develop a series of methods to automatically tag tokens belonging to these classes of ambiguous translations and propose the \textbf{Mu}ltilingual \textbf{D}iscourse-\textbf{A}ware (MuDA) benchmark for context-aware MT models.

%After identifying these linguistic phenomena where context is useful to resolve ambiguity during translation, we develop a series of methods to automatically tag tokens belonging to these classes of ambiguous translations and propose the \textbf{Mu}ltilingual \textbf{D}iscourse-\textbf{A}ware (MuDA) benchmark for context-aware MT models.

\subsection{MT Evaluation Framework}

%\gn{Explain the framework for evaluating MT systems using tagged phenomena}

Given a pair of parallel source and target documents $(X, Y)$, our MuDA tagger assigns one or more tags from a set of discourse phenomena $\{t_i^1, \cdots, t_i^n\}$ to each target token $y_i \in Y$. Using the \texttt{compare-mt} toolkit \cite{neubig-etal-2019-compare}, we compute the mean word f-measure of system outputs compared to the reference for each tag. This allows us to identify which discourse phenomena models can translate more or less accurately.

\begin{figure}[!ht]
    \includegraphics[width=\linewidth]{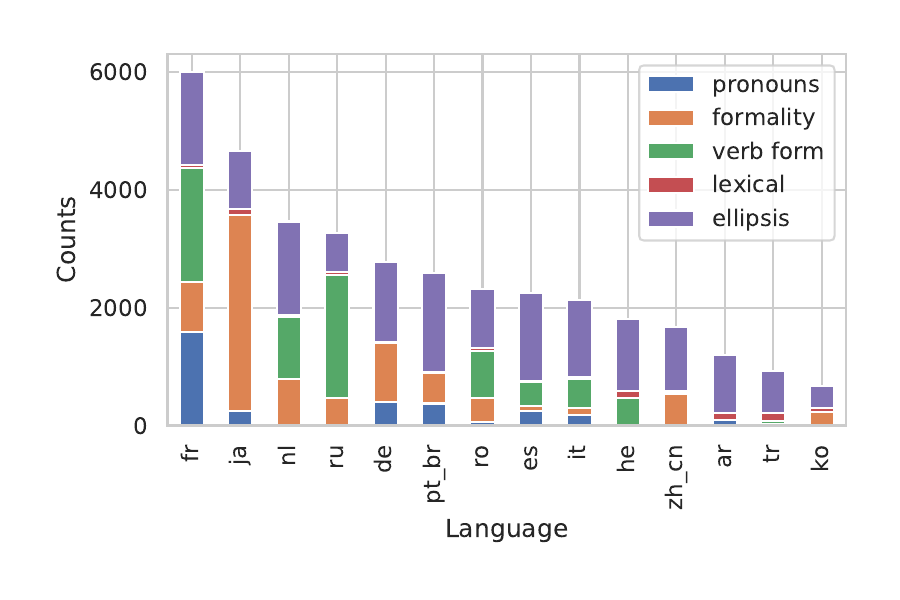}
    \vspace{-2.5em}
    \caption{Number of MuDA tags on TED test data. Exact numbers of each tag are given in \autoref{appendix:total-counts}. Number of tags for other document-level datasets can be found in \autoref{appendix:other-datasets-tags}.}
    \vspace{-1em}
    \label{fig:count_tags}
\end{figure}

\subsection{Automatic Tagging}

We now describe our taggers for each identified discourse phenomenon. Note that these do \textbf{not} require C-XMI to be calculated, and are based on reliable methods for identifying each phenomenon mentioned in \autoref{section:analysis-results}. For formality, pronoun choice and verb form, we created language-specific word lists that were verified by native speakers. 

Not all phenomena are present in each language. Phenomena that are absent are indicated in \autoref{appendix:total-counts}, as a zero count for that language.

%\el{Are all discourse phenomena tagged for all languages? I remember seeing some functions commented out/not existing for languages, we should be clear which phenomena are tagged for which languages if this is the case} \ky{for some languages the phenomena we look at don't appear, which you can see with the "0" number of tags in Table \ref{table:count_tags} (for example, mandarin doesn't have pronoun ambiguity wrt English - which is also described in the appendix. We can make it more explicit which languages don't have some tags if needed}
%and these tags are only applicable to certain target languages that contain the associated discourse phenomenon.

%In doing so, we create more reliable and informative taggers for each phenomenon, rather than using P-CXMI directly to identify ambiguous words, as P-CXMI is fairly noisy and uninterpretable. 

\paragraph{Lexical Cohesion}
To tag words that require lexical cohesion, we first extract word alignments from a parallel corpus $D = \{(X_1, Y_1), \cdots, (X_{|D|},Y_{|D|})\}$, where $(X_m,Y_m)$ denote the source and target reference document pair.
We use the AWESOME aligner \cite{dou-neubig-2021-word} to obtain:
\[
A_m=\{ \langle x_i, y_j\rangle \mid x_i \leftrightarrow y_j, x_i \in X_m, y_j \in Y_m \},
\]
where each $x_i$ and $y_j$ are the lemmatized content source and target words and $\leftrightarrow$ denotes a bidirectional word alignment. 
% Then, for each sentence pair $(\xv_k \in X_m,\yv_k \in Y_m) $, we retrieve:
% \[
% \hat{A}_{m_k} =\{ \langle x_i, y_j\rangle: c(x_i, y_j) | x_i \in \xv_{<k}, y_j \in \yv_{<k} \}
% \]
% \andre{do you mean $\hat{A}_{m,k} =\{ \langle x_i, y_j\rangle \in A_m \mid c_{<k}(x_i, y_j)>3  \}$? maybe it's easier to explain in words}
% where $c(x_i, y_j)$ is the number of occurrences of $\langle x_i, y_j\rangle$ in the previous sentences of the current document.
For each target word $y_j$ that is aligned to source word $x_i$, if the alignment pair $\langle x_i, y_j\rangle$ occurred at least $3$ times already in the current document, excluding the current sentence, we tag $y_j$ for lexical cohesion \footnote{This threshold of 3 can also be changed within the tagger.}.

% \noindent \textbf{Word Sense Disambiguation}
% English has several polysemous words whose meaning and translation depend on context. Moreover, the division in semantic space varies across languages: an English word may have only one possible translation in one language while having multiple non-interchangeable possible translations in another. Therefore, we extract a list of polysemous English words with respect to each target language, and tag all target words that are aligned to a polysemous source word. We identify polysemous source words by first using word alignments to find all possible translations of each content word, and count the occurrence of each translation in our corpus. If the source word appears $n$ times in our corpus, has $m$ different translations and has at least two translations that occur more than $n / m$ times, then we mark this source word as polysemous for the language pair.

\paragraph{Formality}
For languages with T-V distinction, we tag the target pronouns containing formality distinction if there has previously been a word pertaining to the same formality level in the same document. 

Some languages such as Spanish often drop the subject pronoun, and T-V distinction is instead reflected in the verb form. For these languages, we use spaCy \cite{spacy2} and Stanza \cite{qi-etal-2020-stanza} to find POS tags and detect verbs with a second-person subject in the source, and conjugated in the second (T) or third (V) person in the target. 

For languages with more complex honorifics systems, such as Japanese, we construct a word list of common honorifics-related words to tag (details in Appendix~\ref{appendix:verbs}).

\paragraph{Pronoun Choice}
To find pronouns in English that have multiple translations,  we manually construct a list $P_\ell=\{ \langle p_s, \textbf{p}_t \rangle \}$ for each language (Appendix \ref{appendix:pronouns}), where each $p_s$ is an English pronoun and $\textbf{p}_t$ the list of possible translations of $p_s$ in the language $\ell$. Then, for each aligned token pair $\langle x_i, y_j \rangle $, if $x_i, y_j$ are both pronouns with $\langle x_i, \textbf{p}_t | y_j \in \textbf{p}_t \rangle \in P_\ell $, and the antecedent of $x_i$ is \emph{not} in current sentence, we tag $y_j$ as an ambiguous pronoun. To obtain antencedents, we use AllenNLP \cite{Gardner2017AllenNLP}'s coreference resolution module. This procedure is similar to \citet{muller-etal-2018-large}.

\paragraph{Verb Form}
For each target language, we define a list $V_\ell = \{ v_1, \cdots, v_k \}$ of verb forms (Appendix \ref{appendix:verbs}) where $v_i \in V_\ell$ if there exists a verb form in English $u_j$ and an alternate verb form $v_k \neq v_i$ in the target language such that an English verb with form $u_j$ may be translated to a target verb with form $v_i$ or $v_k$ depending on the context. Then, for each target token $y_j$, if $y_j$ is a verb of form  $v_j \in V_\ell$, and another verb with form $v_j$ has appeared previously in the same document, we tag $y_j$ as ambiguous. 

\paragraph{Ellipsis}
To detect translation ambiguity due to VP and NP ellipsis, we look for instances where the ellipsis occurs on the source side, but not on the target side, which means that the ellipsis must be resolved during translation. Since existing ellipsis models are limited to specific types of ellipsis, we first train an English (source-side) ellipsis detection model. To do so, we extract an ellipsis dataset from the English data in the Penn Treebank \citep{marcus-etal-1993-building} and train a BERT text classification model \cite{devlin2018bert}, which achieves 0.77 precision and 0.73 recall (see Appendix \ref{appendix:ellipsis} for training details). 
Then, for each sentence pair where the source sentence is predicted to contain an ellipsis, we tag the word $y_j$ in the target sentence $Y_m$ if: (1) $y_j$ is a verb, noun, proper noun or pronoun; (2) $y_j$ has occurred in the previous target sentences of the same document; (3) $y_j$ is not aligned to any source words, that is, $\not\exists \, x_i \in X_m$ s.t. $\langle x_i, y_j\rangle \in A_m$. 

% \gn{We may want to motivate why we don't use automatic ellipsis detectors (because there are few available, particularly cross-lingually).}
% English text often omits one or more words from a clause when they can be understood from context: for example, in Table \ref{table:high-cxmi}, the verb \textit{die} is elided in the sentence \textit{Those that don't [die]}. However, depending on the nature of the ellipsis and the target language, ellipsis must sometimes be resolved during translation. Although various types of ellipsis exist, we limit our study to verb-phrase (VP) ellipsis and leave the others to future work. To detect when the source text has VP ellipsis but the target text does not, we tag all verbs in the reference text that are not aligned to a source word if the source sentence contains an auxiliary verb or the particle `to'.

% \begin{figure}[t]
%     \includegraphics[width=\linewidth]{count_tags.pdf}
%     \vspace{-6mm}
%     \caption{Number of MuDA tags on TED test data.}
%     \label{fig:count_tags}
%     \vspace{-0.3cm}
% \end{figure}

\subsection{Evaluation of Automatic Tags} 

We apply the MuDA tagger to the reference translations of our TED talk data. We thus obtain an evaluation set of 3,385 parallel sentences for each of the 14 language pairs. In Appendix \ref{appendix:all_cxmi} we report the mean P-CXMI for each language and MuDA tag. Overall, we find higher P-CXMI on tokens with a tag compared to those without, which provides empirical evidence that models indeed rely on context to predict words with MuDA tags. 

%The tags that obtain the highest P-CXMI depends on the target language, hinting that different language pairs contain different translation ambiguities. 

\autoref{appendix:total-counts} shows that the frequency of tags varies significantly across languages. Overall, only 4.5\% of the English sentences have been marked for ellipsis, giving an upper bound for the number of ellipsis tags in other languages. %We suggest our tagger to be applied on a large evaluation set to contain enough examples of ellipsis. 
We find that languages from a different family than English have a relatively high number of ellipsis tags. We also find that Korean and especially Japanese have more formality tags than languages with T-V distinction, which reflects that register is more often important when translating to languages with honorifics. 

\begin{table}[t]
\centering
\resizebox{.9\linewidth}{!}{ 
\begin{tabular}{c|ccccc}
\toprule
 & lexical & formality & pronouns  & verb form & ellipsis\\ 
\midrule
es &  1.00 & 0.92 & 1.00 & 1.00 & 0.53 \\
fr & 1.00 & 1.00 & 1.00  & 0.94 & 0.43 \\
ja & 1.00 & 1.00 & 1.00 & \textendash & 0.41 \\
ko &  1.00 & 0.94 & \textendash & \textendash & 0.26 \\
pt & 0.99 & 0.88 & 1.00 & \textendash & 0.31 \\
ru &  1.00 & 1.00 & \textendash & 1.00 & 0.50 \\
tr & 1.00 & 1.00 & \textendash & 1.00 & 0.57 \\
zh &  1.00 & 1.00 & \textendash & \textendash & 0.78 \\
\bottomrule
\end{tabular}}
\vspace{-1mm}
\caption{Precision of MuDA tags on 50 utterances.} 
\vspace{-3mm}
\label{table:manual}
\end{table}

\paragraph{Manual Evaluation}
%\el{Lack of recall is a bit troublesome here.  Maybe we could also have native speakers find the first 50 examples of phenomena in the shuffled dataset and see the recall of the tagger on the selected examples?}

To evaluate our tagger, we asked native speakers with computational linguistics backgrounds to manually verify MuDA tags for 8 languages on 50 randomly selected utterances as well as all words tagged with \textit{ellipsis} in our corpus. This allows us to measure how many automatic tags violate the given definition of the linguistic tag. Table \ref{table:manual} reports the tags' precision \footnote{Workers were paid 20\$/hour.}. 

For all languages, we obtain high precision for all tags except \textit{ellipsis}, confirming that the methodology can scale to languages where no native speakers were involved in developing the tags. For \textit{ellipsis}, false positives often come from one-to-many or non-literal translations, where the aligner does not align all target words to the corresponding source word. %  Another common mistake is when the reference is not a word-by-word translation of the source expression, so many target words are not aligned to any source words even if there was no ellipsis. 
We believe that the \textit{ellipsis} tagger is still useful in selecting difficult examples that require context for translation; despite the low precision, we find a significantly higher P-CXMI on \textit{ellipsis} words for many languages (Appendix \ref{appendix:all_cxmi}).\footnote{Also note that wrongly assigned tags should also not penalize a system greatly as it should give a low score only if the translation does not match the falsely tagged word.}

\subsection{Extension to New Languages}
\label{sec:extension}
While MuDA currently supports 14 language pairs, our methodology can be easily extended to new languages. The \textit{lexical} and \textit{ellipsis} tags can be directly applied to other languages provided a word aligner between English and the new target language. The \textit{formality} tag can be extended by adding a list of pronouns or verb forms related to formality in the new language. Similarly, the \textit{pronouns} and \textit{verb forms} tag can also be extended by providing a list of ambiguous pronouns and verb forms.

Exhaustively listing all relevant phenomena in document-level MT is extremely complex and beyond the scope of our paper. To identify new discourse phenomena on other languages, our thematic analysis can be reused as follows: (1) Train a model with dynamic context size on translation between the new language pair; (2) Use the model to compute P-CXMI for words in a parallel document-level corpus of the language pair; (3) Manually analyze the POS tags, vocabulary items and individual tokens with high P-CXMI; (4) Link patterns of tokens with high P-CXMI to particular discourse phenomena by consulting linguistic resources.

\begin{figure*}
    \centering
    \includegraphics[width=0.9\textwidth]{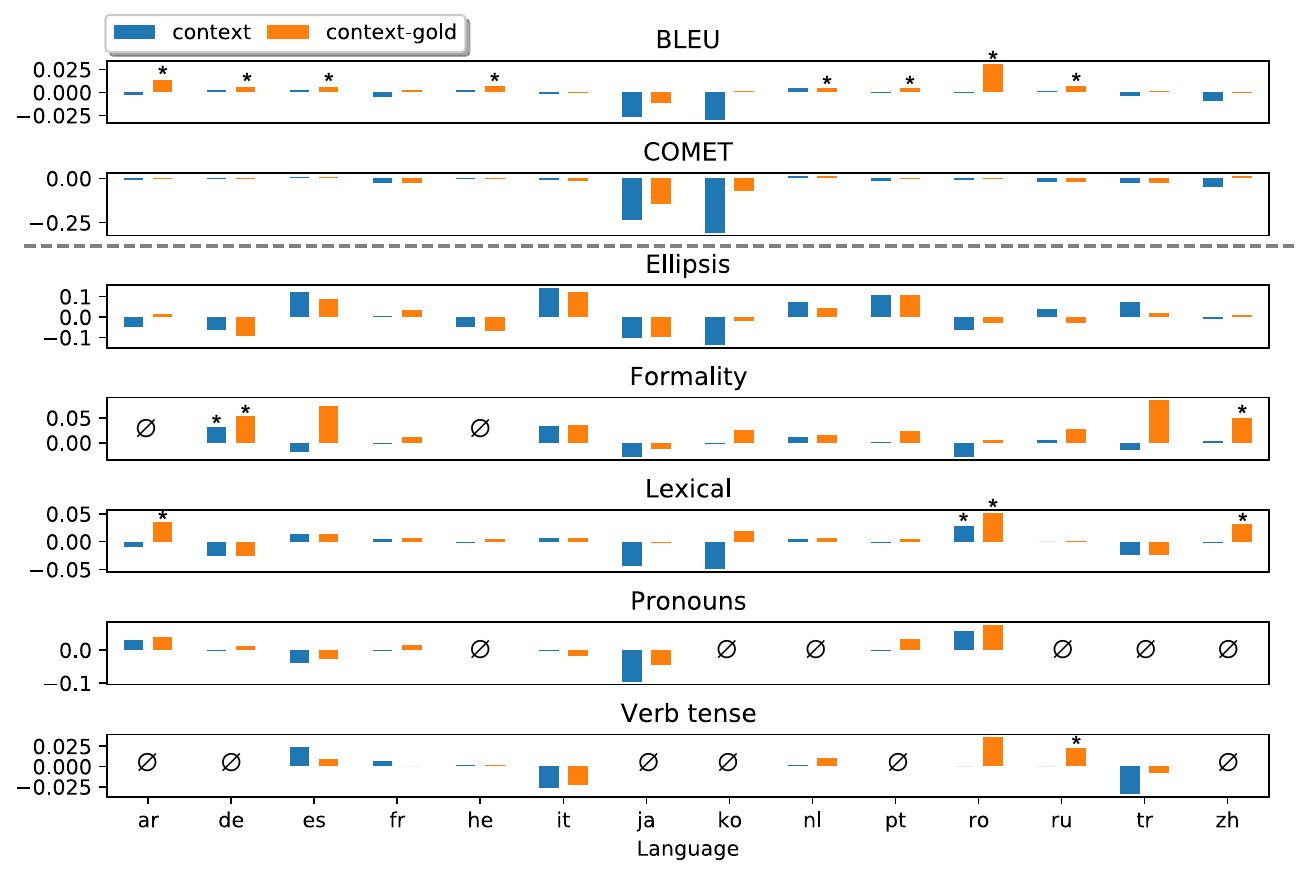}
    \vspace{-0.25cm}
    \caption{Impact of context on BLEU, COMET, and Word f-measure per tag for \texttt{base} context-aware models. BLEU, COMET and word f-measures statistically significantly higher than no-context ($p$ < 0.05) are marked with *. Languages for which the phenomenon doesn't exist are marked with $\emptyset$. BLEU scores are normalized between [0,1]}
    \label{fig:fmeas}
\end{figure*}

\begin{figure}[h!]
    \centering
    \includegraphics[width=0.95\columnwidth]{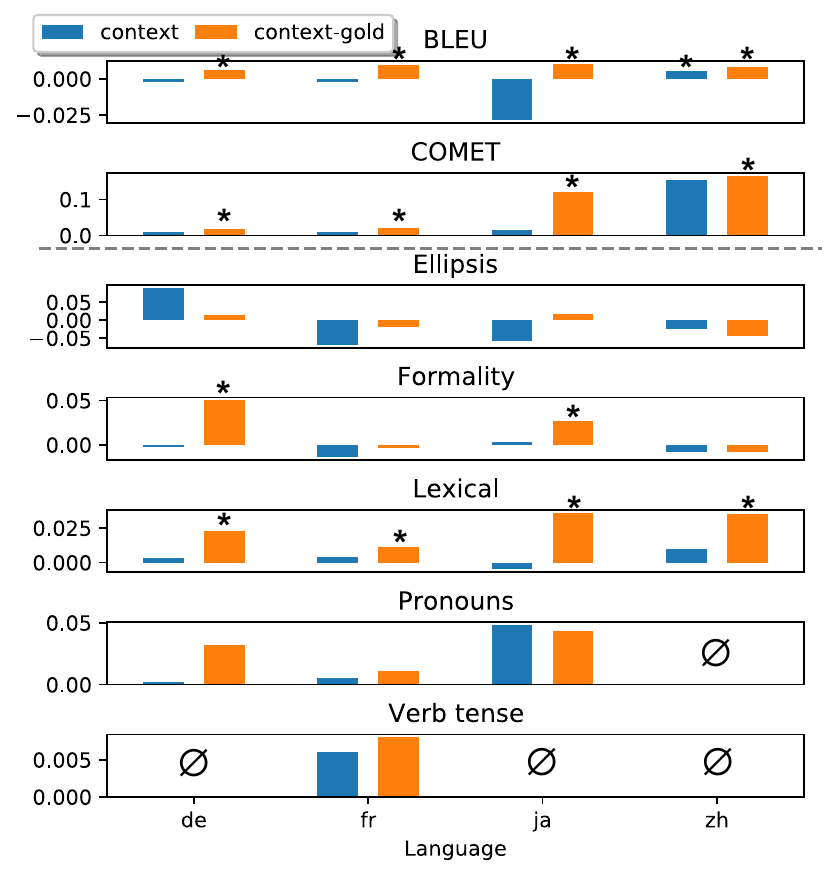}
    \caption{Impact of context on BLEU, COMET, and Word f-measure per tag for \texttt{large} models. Values that are statistically significantly higher than no-context ($p$ < 0.05) are marked with *. Languages for which the phenomenon doesn't exist are marked with $\emptyset$. BLEU scores are normalized between [0,1]}
    \vspace{-0.25cm}
    \label{fig:pre-fmeas}
\end{figure}

\begin{figure*}[h]
    \centering
    \includegraphics[width=0.9\textwidth]{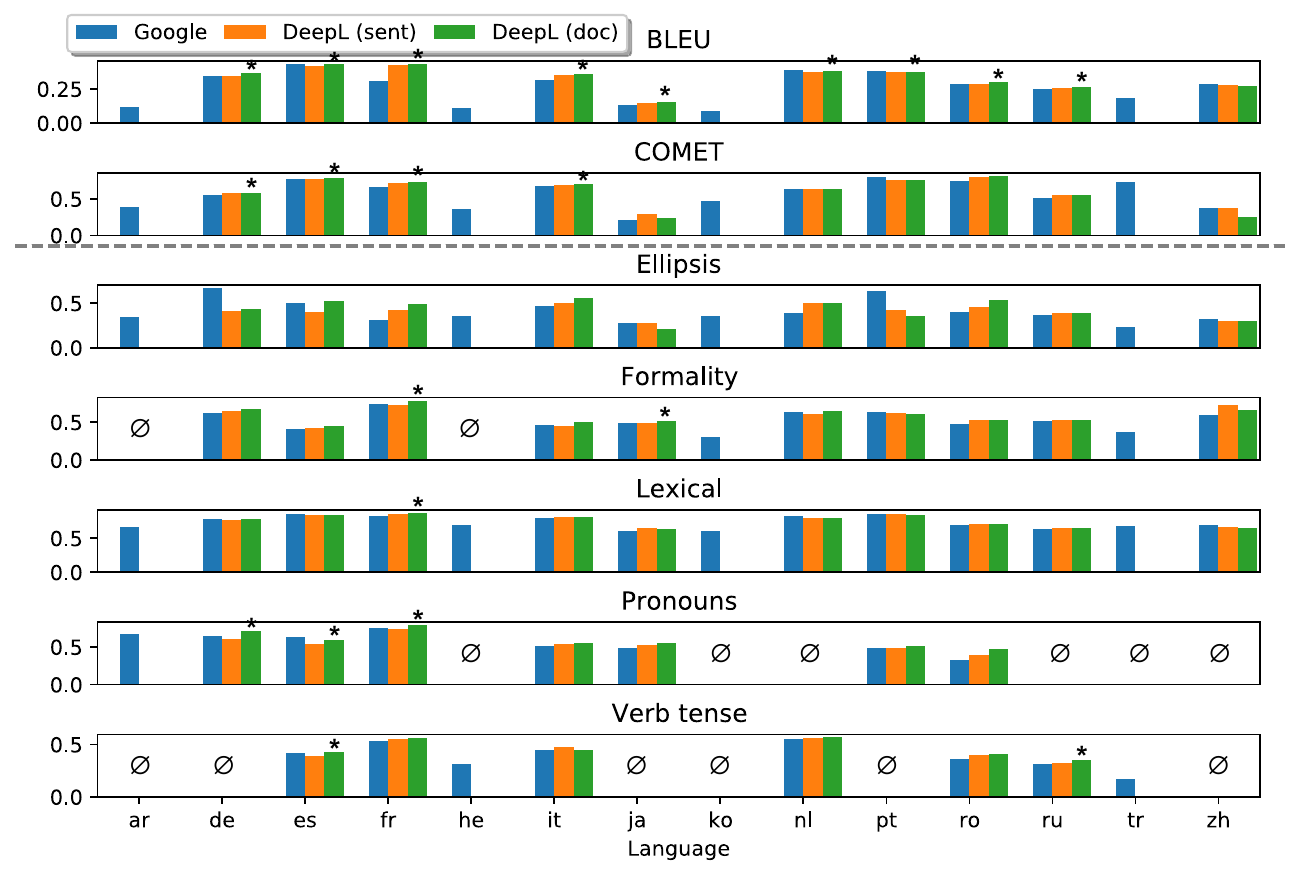}
    \caption{Scores for commercial models. DeepL (doc) BLEU, COMET and word f-measures statistically significantly higher than DeepL (sent) are marked with *. Languages for which neither DeepL or Google translations are available are marked with $\emptyset$. BLEU scores are normalized between [0,1]}
    \label{fig:prov-fmeas}
\end{figure*}

\section{Exploring Context-aware MT}
\label{section:benchmark}

Our MuDA tagger can be applied to documents in the supported languages to create benchmarking datasets for discourse phenomena during translation. We use our benchmark of the TED talk dataset enhanced with MuDA tags to perform an exploration of context usage across languages with 4 models, including commercial systems.

\subsection{Trained Models}

We train a sentence-level and document-level concatenation-based small transformer (\texttt{base}) for every target language.  While conceptually simple, concatenation approaches have been shown to outperform more complex models when properly trained. For the context-aware model, the major difference from \S\ref{sec:data-model-thematic} is that we use a \textit{static} context size of 3, since we are not using these models to measure P-CXMI. \cite{lopes-etal-2020-document}.

To evaluate stronger models, we additionally train a large transformer model (\texttt{large}) that was pretrained on a large, sentence-level corpora, for German, French, Japanese and Chinese. Further details can be found in Appendix~\ref{appendix:training_details}.
%and a \texttt{multi} model - amulti-encoder transformer \andre{typo?} \cite{jean2017does} based on the \textit{transformer-small} architecture that incorporates context by using an aditional encoder for it. \patrick{Should we include info on why we are having these extra models and why these LPs} \andre{yes, maybe a brief explanation?}

% \subsection{Training}

\subsection{Commercial Models}

To assess if commercially available machine translation engines are able to leverage context and therefore do well in MuDA, we consider two engines:\footnote{\url{translate.google.com}, \url{deepl.com}. Translations were obtain from version of engines available in \textbf{April 2021}} (1) the \textit{Google Cloud Translation} v2 API. In early experiments, we assessed that this model only does sentence-level translation, but included it due to its widespread usage; (2) the \textit{DeepL} v2 API. This model advertises its usage of context as part of translations and our experiments confirm this. Early experimentation with other providers (Amazon and Azure) indicated that these are not context-aware so we refrained from evaluating them.

To obtain provider translations, we feed the documents into an API request. To re-segment the translation into sentences, we include special marker tokens in the source that are preserved during translation and split the translation on those tokens. We also evaluate a \textit{sentence-level} version of DeepL where we feed each sentence separately to compare with its document-level counterpart.

\subsection{Results and Discussion}

%\el{The BLEU and COMET scores reported are just for the sentences included in MuDA and not for the overall TED talks right?}
%\ky{BLEU and COMET are over all sentences since we wanted to compare evaluation using corpus-level metrics (without using any info from MuDA) and the targeted evaluation with MuDA, so we didn't filter sentences for BLEU/COMET}
Figure \ref{fig:fmeas} shows results for \texttt{base} models, trained either without (\texttt{no-context}) or with context, and for the latter with either \textit{predicted} (\texttt{context}) or \textit{reference} context (\texttt{context-gold}) during decoding.
Results are reported with respect to standard MT metrics BLEU \cite{papineni-etal-2002-bleu} and COMET \cite{rei-etal-2020-comet}, as well as the MuDA benchmark. The corpus-level metrics BLEU and COMET are calculated over the entire corpus, rather than just the sentences tagged by MuDA. 
%This is in order to compare broad metrics without any additional information given by MuDA with the more fine-grained analysis that MuDA enables.

First, we find that BLEU scores are highest for \texttt{context-gold} models for most language pairs, but context-agnostic models have higher COMET scores. Moreover, in terms of mean word f-measure overall, we do not find significant differences between the three systems. It is therefore difficult to see which system performs the best on document-level ambiguities using only corpus-level metrics.

%\el{It's really hard to scan the chart for this since a lot of the scores aren't statistically significant. E.g. the context-gold models don't seem to have a significant difference from no-context in the ellipsis row?}
For words tagged by MuDA as requiring context for translation, context-aware models often achieve significantly higher word f-measure than context-agnostic models on certain tags such as \textit{ellipsis} and \textit{formality}, but not on other tags such as \textit{lexical} and \textit{verb form}. This demonstrates how MuDA allows us to clarify which inter-sentential ambiguities context-aware models are able to resolve. 
%We do find some phenomena where certain context-aware models do perform better than models without context

For the pretrained \texttt{large} models (Figure  \ref{fig:pre-fmeas}), context-aware models perform better than the context-agnostic on corpus-level metrics, especially COMET. On words tagged with MuDA, context-aware models generally obtain the highest f-measure as well, particularly when given reference context, especially on phenomena such as \textit{lexical} and \textit{pronouns}, but improvements are less pronounced than on corpus-level evaluation.
%For certain tags such as formality, the domain of our data may cause the majority of sentences to be in the formal register, so the baseline model trained on our data may have learned to output the formal register and therefore perform well on the targeted evaluation despite not having contextual information. We therefore posit that models should be trained on open-domain data for MuDA to be able to assess model performance on inter-sentential ambiguities more accurately.

Among commercial engines (Figure \ref{fig:prov-fmeas}), DeepL outperforms Google on most metrics and language pairs. The sentence-level ablation of DeepL performs worse than its document-level system for most MuDA tags.
%, which further suggests DeepL is able to process context to some extent. 

Current context-aware MT systems translate some inter-sentential discourse phenomena well, but are unable to consistently obtain significant improvements over context-agnostic counterparts on challenging MuDA data. Tables with all results can be found in \autoref{appendix:results_tables}.

% Finally, although our \textit{ellipsis} tagger is the least precise tagger in MuDA, our results confirm that it is still effective in evaluating performance in translating context-aware phenomena, as we find that our models and commercial systems do often find some improvements on \textit{ellipsis} with context.

%Nevertheless, evaluation was performed on TED talks data, which is a small, domain-specific dataset. It is therefore unclear the prevalence and importance of discourse phenomena in this data. However, MuDA is dataset agnostic and can be applied to \textit{any} parallel corpora of English and one of our 14 target languages, and can also be extended to more target languages.

\section{Related Work}

%Because examples requiring context to translate accurately is sparse in document-level datasets, standard MT metrics such as BLEU or COMET are not sensitive enough to gains in context-aware MT. Thus, s
%To target evaluation on discourse phenomena, several works resort to measuring the performance of context-aware models targeted to discourse phenomena that require context.

%Some works have attempted to do an automatic evaluation of discourse without relying on a contrastive dataset. 
Several works have worked on measuring the performance of MT models on contextual discourse phenomena. The first example of this was done by \citet{statmt_pronouns}, which evaluated automatically the precision and recall of pronoun translation in statistical MT systems. \citet{jwalapuram-etal-2019-evaluating} proposed evaluating models on pronoun translation based on a pairwise comparison between translations that were generated with and without context, and later \citet{DBLP:journals/corr/abs-2004-14607} extended this work to include more languages and phenomena in their automatic evaluation/test set creation. These works rely on prior domain knowledge and intuition to identify context-aware phenomena, whereas we take a systematic, data-driven approach.
%and find additional phenomena in doing so. 
%Previous approaches also rely on existing systems' outputs and language-specific neural models to generate their evaluation data, whereas MuDA is independent of other systems and do not require additional language-specific data or models, which enables us to study a wider variety of language pairs. 

Most works have focused on evaluating performance in discourse phenomena through the use of \textit{contrastive datasets}.  \citet{muller-etal-2018-large} automatically create a dataset for anaphoric pronoun resolution to evaluate MT models in $\text{EN}\rightarrow \text{DE}$. \citet{bawden-etal-2018-evaluating} manually creates a dataset for both pronoun resolution and lexical choice in $\text{EN}\rightarrow \text{FR}$. \citet{voita-etal-2018-context, voita-etal-2019-good} creates a dataset for anaphora resolution, deixis, ellipsis and lexical cohesion in $\text{EN}\rightarrow\text{RU}$. However, \citet{yin21acl} suggest that \textit{translating} and \textit{disambiguating} between two contrastive choices are inherently different, motivating our approach in measuring direct translation performance. 
\section{Conclusions and Future Work}

We investigate types of ambiguous translations where MT models benefit from context using our proposed P-CXMI metric. We perform a data-driven thematic analysis across 14 languages to identify context-sensitive discourse phenomena, some of which (such as \textit{verb forms}) have not been previously addressed in work on MT. In comparison to previous work, our approach is systematic, extensible, and does not require prior knowledge of the language. Additionally, the P-CXMI metric can be used to identify other context-dependent words in generation.
%The advantages of our approach is that it is systematic and does not require a-priori language-specific knowledge to identify these phenomena, so we believe that our methodology can be easily extended to other language pairs. P-CXMI can also be used to identify types of context-dependent words for tasks outside MT. 
We construct the MuDA benchmark that tags words in parallel corpora and evaluates models on 5 context-dependent phenomena. Our evaluation reveals that context-aware and commercial translation systems achieve small improvements over context-agnostic models on our benchmark, and we encourage further development of models that improve on context-aware translation. 
%We find that \textit{ellipsis} is the most challenging to tag with high precision and we leave improvements to model cross-lingual ellipsis for future work.

\section*{Limitations}

While MuDA relies on set of hand-crafted rules for tagging specific phenomena, these rules might involve the use of other error-prone systems (such as coreference resolution and alignment models) and these errors might be susceptible to problems (such as lack of out-of-domain generalization) that could limit the applicability of our tagger. However, this could be fixed by extending MuDA to use newer and better versions of these systems.

The use of F-1 per tag with surface-form matching between reference/translation can also lead to penalizing translations that use context correctly but choose other equivalent words. Nevertheless, this should also be mitigable by extending the scoring method to, for example, match synonyms.

Finally, the benchmarking of context-aware models might not apply to newer, state-of-the-art translation models, especially if these leverage large language models that were trained on long-context data.

\section*{Acknowledgements}

% \andre{this needs to be removed!!!}
We would like to thank Uri Alon, Ipek Baris, George Bejinariu, Hiba Belkadi, Chloé Billiotte, Giovanni Campagna, Remi Castera, Volkan Cirik, Taisiya Glushkova, Junxian He, Mert Inan, Alina Karakanta, Benno Krojer, Emma Landry, Chanyoung Park, Artidoro Pagnoni, Maria Ryskina, Odette Scharenborg, Melanie Sclar, Jenny Seok, Emma Schippers, Bogdan Vasilescu for advice on various languages and help with manual annotations.

We would also like to thank all the members of DeepSPIN and NeuLab who provided feedback on earlier versions of this work. This work was  supported by the European Research Council (ERC StG DeepSPIN 758969), by EU's Horizon Europe Research and Innovation Actions (UTTER, contract 101070631), by the P2020 program MAIA (LISBOA-01-0247-FEDER-045909), by the Portuguese Recovery and Resilience Plan  through project C645008882-00000055 (NextGenAI, Center for Responsible AI), and by the Funda\c{c}\~ao para a Ci\^encia e Tecnologia through contracts SFRH/BD/150706/2020 and UIDB/50008/2020.

\bibliographystyle{acl_natbib}
\bibliography{anthology,emnlp2021}

\begin{thebibliography}{39}
\expandafter\ifx\csname natexlab\endcsname\relax\def\natexlab#1{#1}\fi

\bibitem[{Barrault et~al.(2019)Barrault, Bojar, Costa-juss{\`a}, Federmann,
  Fishel, Graham, Haddow, Huck, Koehn, Malmasi, Monz, M{\"u}ller, Pal, Post,
  and Zampieri}]{news-commentary}
Lo{\"\i}c Barrault, Ond{\v{r}}ej Bojar, Marta~R. Costa-juss{\`a}, Christian
  Federmann, Mark Fishel, Yvette Graham, Barry Haddow, Matthias Huck, Philipp
  Koehn, Shervin Malmasi, Christof Monz, Mathias M{\"u}ller, Santanu Pal, Matt
  Post, and Marcos Zampieri. 2019.
\newblock \href {https://doi.org/10.18653/v1/W19-5301} {Findings of the 2019
  conference on machine translation ({WMT}19)}.
\newblock In \emph{Proceedings of the Fourth Conference on Machine Translation
  (Volume 2: Shared Task Papers, Day 1)}, pages 1--61, Florence, Italy.
  Association for Computational Linguistics.

\bibitem[{Bawden et~al.(2018)Bawden, Sennrich, Birch, and
  Haddow}]{bawden-etal-2018-evaluating}
Rachel Bawden, Rico Sennrich, Alexandra Birch, and Barry Haddow. 2018.
\newblock \href {https://doi.org/10.18653/v1/N18-1118} {Evaluating discourse
  phenomena in neural machine translation}.
\newblock In \emph{Proceedings of the 2018 Conference of the North {A}merican
  Chapter of the Association for Computational Linguistics: Human Language
  Technologies, Volume 1 (Long Papers)}, pages 1304--1313, New Orleans,
  Louisiana. Association for Computational Linguistics.

\bibitem[{Bies et~al.(1995)Bies, Ferguson, Katz, MacIntyre, Tredinnick, Kim,
  Marcinkiewicz, and Schasberger}]{bies1995bracketing}
Ann Bies, Mark Ferguson, Karen Katz, Robert MacIntyre, Victoria Tredinnick,
  Grace Kim, Mary~Ann Marcinkiewicz, and Britta Schasberger. 1995.
\newblock Bracketing guidelines for treebank ii style penn treebank project.
\newblock \emph{University of Pennsylvania}, 97:100.

\bibitem[{Braun and Clarke(2006)}]{braun2006using}
Virginia Braun and Victoria Clarke. 2006.
\newblock Using thematic analysis in psychology.
\newblock \emph{Qualitative research in psychology}, 3(2):77--101.

\bibitem[{Bugliarello et~al.(2020)Bugliarello, Mielke, Anastasopoulos,
  Cotterell, and Okazaki}]{bugliarello-etal-2020-easier}
Emanuele Bugliarello, Sabrina~J. Mielke, Antonios Anastasopoulos, Ryan
  Cotterell, and Naoaki Okazaki. 2020.
\newblock \href {https://doi.org/10.18653/v1/2020.acl-main.149} {It{'}s easier
  to translate out of {E}nglish than into it: {M}easuring neural translation
  difficulty by cross-mutual information}.
\newblock In \emph{Proceedings of the 58th Annual Meeting of the Association
  for Computational Linguistics}, pages 1640--1649, Online. Association for
  Computational Linguistics.

\bibitem[{Cettolo et~al.(2012)Cettolo, Girardi, and Federico}]{cettolo-2012}
Mauro Cettolo, Christian Girardi, and Marcello Federico. 2012.
\newblock \href {https://www.aclweb.org/anthology/2012.eamt-1.60} {{WIT}3: Web
  inventory of transcribed and translated talks}.
\newblock In \emph{Proceedings of the 16th Annual conference of the European
  Association for Machine Translation}, pages 261--268, Trento, Italy. European
  Association for Machine Translation.

\bibitem[{Church and Hanks(1990)}]{church-hanks-1990-word}
Kenneth~Ward Church and Patrick Hanks. 1990.
\newblock \href {https://www.aclweb.org/anthology/J90-1003} {Word association
  norms, mutual information, and lexicography}.
\newblock \emph{Computational Linguistics}, 16(1):22--29.

\bibitem[{Devlin et~al.(2019)Devlin, Chang, Lee, and
  Toutanova}]{devlin2018bert}
Jacob Devlin, Ming-Wei Chang, Kenton Lee, and Kristina Toutanova. 2019.
\newblock \href {https://doi.org/10.18653/v1/N19-1423} {{BERT}: Pre-training of
  deep bidirectional transformers for language understanding}.
\newblock In \emph{Proceedings of the 2019 Conference of the North {A}merican
  Chapter of the Association for Computational Linguistics: Human Language
  Technologies, Volume 1 (Long and Short Papers)}, pages 4171--4186,
  Minneapolis, Minnesota. Association for Computational Linguistics.

\bibitem[{Dou and Neubig(2021)}]{dou-neubig-2021-word}
Zi-Yi Dou and Graham Neubig. 2021.
\newblock \href {https://www.aclweb.org/anthology/2021.eacl-main.181} {Word
  alignment by fine-tuning embeddings on parallel corpora}.
\newblock In \emph{Proceedings of the 16th Conference of the European Chapter
  of the Association for Computational Linguistics: Main Volume}, pages
  2112--2128, Online. Association for Computational Linguistics.

\bibitem[{Espl{\`a} et~al.(2019)Espl{\`a}, Forcada, Ram{\'\i}rez-S{\'a}nchez,
  and Hoang}]{espla-etal-2019-paracrawl}
Miquel Espl{\`a}, Mikel Forcada, Gema Ram{\'\i}rez-S{\'a}nchez, and Hieu Hoang.
  2019.
\newblock \href {https://www.aclweb.org/anthology/W19-6721} {{P}ara{C}rawl:
  Web-scale parallel corpora for the languages of the {EU}}.
\newblock In \emph{Proceedings of Machine Translation Summit XVII Volume 2:
  Translator, Project and User Tracks}, pages 118--119, Dublin, Ireland.
  European Association for Machine Translation.

\bibitem[{Fernandes et~al.(2021)Fernandes, Yin, Neubig, and
  Martins}]{fernandes21acl}
Patrick Fernandes, Kayo Yin, Graham Neubig, and André F.~T. Martins. 2021.
\newblock \href {https://arxiv.org/abs/2105.03482} {Measuring and increasing
  context usage in context-aware machine translation}.
\newblock In \emph{Joint Conference of the 59th Annual Meeting of the
  Association for Computational Linguistics and the 11th International Joint
  Conference on Natural Language Processing (ACL-IJCNLP)}, Virtual.

\bibitem[{Gardner et~al.(2017)Gardner, Grus, Neumann, Tafjord, Dasigi, Liu,
  Peters, Schmitz, and Zettlemoyer}]{Gardner2017AllenNLP}
Matt Gardner, Joel Grus, Mark Neumann, Oyvind Tafjord, Pradeep Dasigi,
  Nelson~F. Liu, Matthew Peters, Michael Schmitz, and Luke~S. Zettlemoyer.
  2017.
\newblock \href {http://arxiv.org/abs/arXiv:1803.07640} {Allennlp: A deep
  semantic natural language processing platform}.

\bibitem[{Guillou et~al.(2018)Guillou, Hardmeier, Lapshinova-Koltunski, and
  Lo{\'a}iciga}]{guillou-etal-2018-pronoun}
Liane Guillou, Christian Hardmeier, Ekaterina Lapshinova-Koltunski, and Sharid
  Lo{\'a}iciga. 2018.
\newblock \href {https://doi.org/10.18653/v1/W18-6435} {A pronoun test suite
  evaluation of the {E}nglish{--}{G}erman {MT} systems at {WMT} 2018}.
\newblock In \emph{Proceedings of the Third Conference on Machine Translation:
  Shared Task Papers}, pages 570--577, Belgium, Brussels. Association for
  Computational Linguistics.

\bibitem[{Hardmeier et~al.(2010)Hardmeier, Fondazione, and
  Kessler}]{statmt_pronouns}
Christian Hardmeier, Marcello Fondazione, and Bruno Kessler. 2010.
\newblock Modelling pronominal anaphora in statistical machine translation.

\bibitem[{Honnibal and Montani(2017)}]{spacy2}
Matthew Honnibal and Ines Montani. 2017.
\newblock {spaCy 2}: Natural language understanding with {B}loom embeddings,
  convolutional neural networks and incremental parsing.
\newblock To appear.

\bibitem[{Jwalapuram et~al.(2019)Jwalapuram, Joty, Temnikova, and
  Nakov}]{jwalapuram-etal-2019-evaluating}
Prathyusha Jwalapuram, Shafiq Joty, Irina Temnikova, and Preslav Nakov. 2019.
\newblock \href {https://doi.org/10.18653/v1/D19-1294} {Evaluating pronominal
  anaphora in machine translation: An evaluation measure and a test suite}.
\newblock In \emph{Proceedings of the 2019 Conference on Empirical Methods in
  Natural Language Processing and the 9th International Joint Conference on
  Natural Language Processing (EMNLP-IJCNLP)}, pages 2964--2975, Hong Kong,
  China. Association for Computational Linguistics.

\bibitem[{Jwalapuram et~al.(2020)Jwalapuram, Rychalska, Joty, and
  Basaj}]{DBLP:journals/corr/abs-2004-14607}
Prathyusha Jwalapuram, Barbara Rychalska, Shafiq~R. Joty, and Dominika Basaj.
  2020.
\newblock \href {http://arxiv.org/abs/2004.14607} {Can your context-aware {MT}
  system pass the dip benchmark tests? : Evaluation benchmarks for discourse
  phenomena in machine translation}.
\newblock \emph{CoRR}, abs/2004.14607.

\bibitem[{L{\"a}ubli et~al.(2018)L{\"a}ubli, Sennrich, and
  Volk}]{laubli-etal-2018-machine}
Samuel L{\"a}ubli, Rico Sennrich, and Martin Volk. 2018.
\newblock \href {https://doi.org/10.18653/v1/D18-1512} {Has machine translation
  achieved human parity? a case for document-level evaluation}.
\newblock In \emph{Proceedings of the 2018 Conference on Empirical Methods in
  Natural Language Processing}, pages 4791--4796, Brussels, Belgium.
  Association for Computational Linguistics.

\bibitem[{Lopes et~al.(2020)Lopes, Farajian, Bawden, Zhang, and
  Martins}]{lopes-etal-2020-document}
Ant{\'o}nio Lopes, M.~Amin Farajian, Rachel Bawden, Michael Zhang, and
  Andr{\'e} F.~T. Martins. 2020.
\newblock \href {https://www.aclweb.org/anthology/2020.eamt-1.24}
  {Document-level neural {MT}: A systematic comparison}.
\newblock In \emph{Proceedings of the 22nd Annual Conference of the European
  Association for Machine Translation}, pages 225--234, Lisboa, Portugal.
  European Association for Machine Translation.

\bibitem[{Marcus et~al.(1993)Marcus, Santorini, and
  Marcinkiewicz}]{marcus-etal-1993-building}
Mitchell~P. Marcus, Beatrice Santorini, and Mary~Ann Marcinkiewicz. 1993.
\newblock \href {https://www.aclweb.org/anthology/J93-2004} {Building a large
  annotated corpus of {E}nglish: The {P}enn {T}reebank}.
\newblock \emph{Computational Linguistics}, 19(2):313--330.

\bibitem[{Maruf and Haffari(2018)}]{maruf-haffari-2018-document}
Sameen Maruf and Gholamreza Haffari. 2018.
\newblock \href {https://doi.org/10.18653/v1/P18-1118} {Document context neural
  machine translation with memory networks}.
\newblock In \emph{Proceedings of the 56th Annual Meeting of the Association
  for Computational Linguistics (Volume 1: Long Papers)}, pages 1275--1284,
  Melbourne, Australia. Association for Computational Linguistics.

\bibitem[{Maruf et~al.(2021)Maruf, Saleh, and Haffari}]{10.1145/3441691}
Sameen Maruf, Fahimeh Saleh, and Gholamreza Haffari. 2021.
\newblock \href {https://doi.org/10.1145/3441691} {A survey on document-level
  neural machine translation: Methods and evaluation}.
\newblock \emph{ACM Comput. Surv.}, 54(2).

\bibitem[{Miculicich et~al.(2018)Miculicich, Ram, Pappas, and
  Henderson}]{miculicich-etal-2018-document}
Lesly Miculicich, Dhananjay Ram, Nikolaos Pappas, and James Henderson. 2018.
\newblock \href {https://doi.org/10.18653/v1/D18-1325} {Document-level neural
  machine translation with hierarchical attention networks}.
\newblock In \emph{Proceedings of the 2018 Conference on Empirical Methods in
  Natural Language Processing}, pages 2947--2954, Brussels, Belgium.
  Association for Computational Linguistics.

\bibitem[{Morishita et~al.(2020)Morishita, Suzuki, and
  Nagata}]{morishita-etal-2020-jparacrawl}
Makoto Morishita, Jun Suzuki, and Masaaki Nagata. 2020.
\newblock \href {https://www.aclweb.org/anthology/2020.lrec-1.443}
  {{JP}ara{C}rawl: A large scale web-based {E}nglish-{J}apanese parallel
  corpus}.
\newblock In \emph{Proceedings of the 12th Language Resources and Evaluation
  Conference}, pages 3603--3609, Marseille, France. European Language Resources
  Association.

\bibitem[{M{\"u}ller et~al.(2018)M{\"u}ller, Rios, Voita, and
  Sennrich}]{muller-etal-2018-large}
Mathias M{\"u}ller, Annette Rios, Elena Voita, and Rico Sennrich. 2018.
\newblock \href {https://doi.org/10.18653/v1/W18-6307} {A large-scale test set
  for the evaluation of context-aware pronoun translation in neural machine
  translation}.
\newblock In \emph{Proceedings of the Third Conference on Machine Translation:
  Research Papers}, pages 61--72, Brussels, Belgium. Association for
  Computational Linguistics.

\bibitem[{Neubig et~al.(2019)Neubig, Dou, Hu, Michel, Pruthi, and
  Wang}]{neubig-etal-2019-compare}
Graham Neubig, Zi-Yi Dou, Junjie Hu, Paul Michel, Danish Pruthi, and Xinyi
  Wang. 2019.
\newblock \href {https://doi.org/10.18653/v1/N19-4007} {compare-mt: A tool for
  holistic comparison of language generation systems}.
\newblock In \emph{Proceedings of the 2019 Conference of the North {A}merican
  Chapter of the Association for Computational Linguistics (Demonstrations)},
  pages 35--41, Minneapolis, Minnesota. Association for Computational
  Linguistics.

\bibitem[{Ott et~al.(2019)Ott, Edunov, Baevski, Fan, Gross, Ng, Grangier, and
  Auli}]{ott-etal-2019-fairseq}
Myle Ott, Sergey Edunov, Alexei Baevski, Angela Fan, Sam Gross, Nathan Ng,
  David Grangier, and Michael Auli. 2019.
\newblock \href {https://doi.org/10.18653/v1/N19-4009} {fairseq: A fast,
  extensible toolkit for sequence modeling}.
\newblock In \emph{Proceedings of the 2019 Conference of the North {A}merican
  Chapter of the Association for Computational Linguistics (Demonstrations)},
  pages 48--53, Minneapolis, Minnesota. Association for Computational
  Linguistics.

\bibitem[{Papineni et~al.(2002)Papineni, Roukos, Ward, and
  Zhu}]{papineni-etal-2002-bleu}
Kishore Papineni, Salim Roukos, Todd Ward, and Wei-Jing Zhu. 2002.
\newblock \href {https://doi.org/10.3115/1073083.1073135} {{B}leu: a method for
  automatic evaluation of machine translation}.
\newblock In \emph{Proceedings of the 40th Annual Meeting of the Association
  for Computational Linguistics}, pages 311--318, Philadelphia, Pennsylvania,
  USA. Association for Computational Linguistics.

\bibitem[{Qi et~al.(2020)Qi, Zhang, Zhang, Bolton, and
  Manning}]{qi-etal-2020-stanza}
Peng Qi, Yuhao Zhang, Yuhui Zhang, Jason Bolton, and Christopher~D. Manning.
  2020.
\newblock \href {https://doi.org/10.18653/v1/2020.acl-demos.14} {{S}tanza: A
  python natural language processing toolkit for many human languages}.
\newblock In \emph{Proceedings of the 58th Annual Meeting of the Association
  for Computational Linguistics: System Demonstrations}, pages 101--108,
  Online. Association for Computational Linguistics.

\bibitem[{Qi et~al.(2018)Qi, Sachan, Felix, Padmanabhan, and
  Neubig}]{qi-etal-2018-pre}
Ye~Qi, Devendra Sachan, Matthieu Felix, Sarguna Padmanabhan, and Graham Neubig.
  2018.
\newblock \href {https://doi.org/10.18653/v1/N18-2084} {When and why are
  pre-trained word embeddings useful for neural machine translation?}
\newblock In \emph{Proceedings of the 2018 Conference of the North {A}merican
  Chapter of the Association for Computational Linguistics: Human Language
  Technologies, Volume 2 (Short Papers)}, pages 529--535, New Orleans,
  Louisiana. Association for Computational Linguistics.

\bibitem[{Rei et~al.(2020)Rei, Stewart, Farinha, and
  Lavie}]{rei-etal-2020-comet}
Ricardo Rei, Craig Stewart, Ana~C Farinha, and Alon Lavie. 2020.
\newblock \href {https://doi.org/10.18653/v1/2020.emnlp-main.213} {{COMET}: A
  neural framework for {MT} evaluation}.
\newblock In \emph{Proceedings of the 2020 Conference on Empirical Methods in
  Natural Language Processing (EMNLP)}, pages 2685--2702, Online. Association
  for Computational Linguistics.

\bibitem[{Tiedemann and Scherrer(2017)}]{tiedemann-scherrer-2017-neural}
J{\"o}rg Tiedemann and Yves Scherrer. 2017.
\newblock \href {https://doi.org/10.18653/v1/W17-4811} {Neural machine
  translation with extended context}.
\newblock In \emph{Proceedings of the Third Workshop on Discourse in Machine
  Translation}, pages 82--92, Copenhagen, Denmark. Association for
  Computational Linguistics.

\bibitem[{Toral et~al.(2018)Toral, Castilho, Hu, and
  Way}]{toral-etal-2018-attaining}
Antonio Toral, Sheila Castilho, Ke~Hu, and Andy Way. 2018.
\newblock \href {https://doi.org/10.18653/v1/W18-6312} {Attaining the
  unattainable? reassessing claims of human parity in neural machine
  translation}.
\newblock In \emph{Proceedings of the Third Conference on Machine Translation:
  Research Papers}, pages 113--123, Brussels, Belgium. Association for
  Computational Linguistics.

\bibitem[{Vaswani et~al.(2017)Vaswani, Shazeer, Parmar, Uszkoreit, Jones,
  Gomez, Kaiser, and Polosukhin}]{46201}
Ashish Vaswani, Noam Shazeer, Niki Parmar, Jakob Uszkoreit, Llion Jones,
  Aidan~N. Gomez, Lukasz Kaiser, and Illia Polosukhin. 2017.
\newblock \href
  {https://proceedings.neurips.cc/paper/2017/hash/3f5ee243547dee91fbd053c1c4a845aa-Abstract.html}
  {Attention is all you need}.
\newblock In \emph{Advances in Neural Information Processing Systems 30: Annual
  Conference on Neural Information Processing Systems 2017, December 4-9, 2017,
  Long Beach, CA, {USA}}, pages 5998--6008.

\bibitem[{Voita et~al.(2019{\natexlab{a}})Voita, Sennrich, and
  Titov}]{voita-etal-2019-context}
Elena Voita, Rico Sennrich, and Ivan Titov. 2019{\natexlab{a}}.
\newblock \href {https://doi.org/10.18653/v1/D19-1081} {Context-aware
  monolingual repair for neural machine translation}.
\newblock In \emph{Proceedings of the 2019 Conference on Empirical Methods in
  Natural Language Processing and the 9th International Joint Conference on
  Natural Language Processing (EMNLP-IJCNLP)}, pages 877--886, Hong Kong,
  China. Association for Computational Linguistics.

\bibitem[{Voita et~al.(2019{\natexlab{b}})Voita, Sennrich, and
  Titov}]{voita-etal-2019-good}
Elena Voita, Rico Sennrich, and Ivan Titov. 2019{\natexlab{b}}.
\newblock \href {https://doi.org/10.18653/v1/P19-1116} {When a good translation
  is wrong in context: Context-aware machine translation improves on deixis,
  ellipsis, and lexical cohesion}.
\newblock In \emph{Proceedings of the 57th Annual Meeting of the Association
  for Computational Linguistics}, pages 1198--1212, Florence, Italy.
  Association for Computational Linguistics.

\bibitem[{Voita et~al.(2018)Voita, Serdyukov, Sennrich, and
  Titov}]{voita-etal-2018-context}
Elena Voita, Pavel Serdyukov, Rico Sennrich, and Ivan Titov. 2018.
\newblock \href {https://doi.org/10.18653/v1/P18-1117} {Context-aware neural
  machine translation learns anaphora resolution}.
\newblock In \emph{Proceedings of the 56th Annual Meeting of the Association
  for Computational Linguistics (Volume 1: Long Papers)}, pages 1264--1274,
  Melbourne, Australia. Association for Computational Linguistics.

\bibitem[{Wolf et~al.(2020)Wolf, Debut, Sanh, Chaumond, Delangue, Moi, Cistac,
  Rault, Louf, Funtowicz, Davison, Shleifer, von Platen, Ma, Jernite, Plu, Xu,
  Le~Scao, Gugger, Drame, Lhoest, and Rush}]{wolf-etal-2020-transformers}
Thomas Wolf, Lysandre Debut, Victor Sanh, Julien Chaumond, Clement Delangue,
  Anthony Moi, Pierric Cistac, Tim Rault, Remi Louf, Morgan Funtowicz, Joe
  Davison, Sam Shleifer, Patrick von Platen, Clara Ma, Yacine Jernite, Julien
  Plu, Canwen Xu, Teven Le~Scao, Sylvain Gugger, Mariama Drame, Quentin Lhoest,
  and Alexander Rush. 2020.
\newblock \href {https://doi.org/10.18653/v1/2020.emnlp-demos.6} {Transformers:
  State-of-the-art natural language processing}.
\newblock In \emph{Proceedings of the 2020 Conference on Empirical Methods in
  Natural Language Processing: System Demonstrations}, pages 38--45, Online.
  Association for Computational Linguistics.

\bibitem[{Yin et~al.(2021)Yin, Fernandes, Pruthi, Chaudhary, Martins, and
  Neubig}]{yin21acl}
Kayo Yin, Patrick Fernandes, Danish Pruthi, Aditi Chaudhary, André F.~T.
  Martins, and Graham Neubig. 2021.
\newblock \href {https://arxiv.org/abs/2105.06977} {Do context-aware
  translation models pay the right attention?}
\newblock In \emph{Joint Conference of the 59th Annual Meeting of the
  Association for Computational Linguistics and the 11th International Joint
  Conference on Natural Language Processing (ACL-IJCNLP)}, Virtual.

\end{thebibliography}

\onecolumn

\clearpage

\newpage

\appendix

\section{MuDA Toolkit Usage}
\label{appendix:muda-toolkit}

\lstset{
  basicstyle=\ttfamily,
  columns=fullflexible,
  frame=single,
  breaklines=true,
  postbreak=\mbox{\textcolor{red}{$\hookrightarrow$}\space},
  showstringspaces=false,
  commentstyle=\color{gray},
  keywordstyle=[1]\color{blue},
  keywordstyle=[2]\color{purple},
  stringstyle=\color{orange},
  numbers=left,
  numberstyle=\tiny\color{gray},
  captionpos=b,
  language=bash
}

To tag an existing dataset and extract the tags for later use, run the following command:

\begin{lstlisting}[language=bash]
python muda/main.py \
    --src /path/to/src \
    --tgt /path/to/tgt \
    --docids /path/to/docids \
    --dump-tags /tmp/maia_ende.tags \
    --tgt-lang lang
\end{lstlisting}

    To evaluate models on a particular dataset (reporting per-tag metrics dicussed in this paper), run the following command:

\begin{lstlisting}[language=bash]
python muda/main.py \
    --src /path/to/src \
    --tgt /path/to/tgt \
    --docids /path/to/docids \
    --hyps /path/to/hyps.m1 /path/to/hyps.m2 \
    --tgt-lang lang
\end{lstlisting}

\section{Language Properties}
\label{appendix:lang}

\begin{table*}[t]
\resizebox{\linewidth}{!}{ 
\begin{tabular}{c|ccccc}
\toprule
Language & Family & Word Order & Pronouns Politeness & Gendered Pronouns & Gender Assignment \\
\midrule
Arabic & Afro-Asiatic & VSO & None & 1 and/or 2 and 3 & Semantic-Formal \\ 
English & Indo-European & SVO & None & 3.Sing & Semantic \\
German & Indo-European & SOV/SVO & Binary & 3.Sing & Semantic-Formal\\
Spanish & Indo-European & SVO & Binary & 1 and/or 2 and 3 & Semantic-Formal\\
French & Indo-European  & SVO & Binary & 3.Sing & Semantic-Formal \\
Hebrew & Afro-Asiatic & SVO & None & 1 and/or 2 and 3 & Semantic-Formal\\
Italian & Indo-European & SVO & Binary & 3.Sing & Semantic-Formal\\
Japanese & Japonic & SOV & Avoided & 3 & None \\
Korean & Koreanic & SOV & Avoided & 3.Sing & None \\
Dutch & Indo-European & SOV/SVO & Binary & 3.Sing & Semantic-Formal \\
Portuguese & Indo-European & SVO & Binary& 3.Sing & Semantic-Formal \\
Romanian & Indo-European & SVO& Multiple &  3.Sing & Semantic-Formal \\
Russian & Indo-European & SVO &  Binary & 3.Sing & Semantic-Formal \\
Turkish & Turkic & SOV & Binary & None & None \\
Mandarin & Sino-Tibetan & SVO & Binary & 3.Sing & None \\
\bottomrule
\end{tabular}}
\vspace{-1mm}
\caption{Properties of the languages in our study. %\ky{TODO: cite Ethnologue, WALS. any other relevant features?} \gn{This could be summarized in the main text and moved to an appendix if we need to save space.} 
} 
\vspace{-2mm}
\label{table:langs}
\end{table*}

Table \ref{table:langs} summarizes the properties of the languages analyzed in this work.

\section{P-CXMI Results}
\label{appendix:all_cxmi}
\begin{table*}[t]
\resizebox{\linewidth}{!}{ 
\begin{tabular}{c|cccccccccccccc}
\toprule
& ar & de &	es&	fr&	he&	it&	ja&	ko&	nl&	pt&	ro&	ru&	tr&	zh \\
\midrule
CXMI & 0.073 & 0.008 & 0.011 & 0.011 & 0.021 & 0.015 & 0.067 & 0.035 & 0.005 & 0.009 & 0.051 & 0.015 & 0.016 & 0.081 \\
P-CXMI & 0.075 & 0.005 & 0.011 & 0.021 & 0.023 & 0.016 & 0.059 & 0.038 & 0.002 & 0.013 & 0.049 & 0.015 & 0.014 & 0.057 \\
\midrule
ADJ & 0.017 & -0.014 & -0.011 & 0.000 & -0.037 & -0.008 & 0.001 & -0.002 & -0.006 & -0.005 & 0.020 & 0.015 & -0.006 & 0.007 \\
ADP & 0.017 & -0.001 & -0.004 & -0.004 & -0.006 & -0.005 & 0.005 & 0.014 & -0.005 & -0.001 & 0.011 & -0.003 & -0.005 & -0.001 \\
ADV & 0.038 & -0.011 & 0.008 & 0.002 & 0.007 & 0.005 & 0.005 & -0.006 & 0.001 & 0.011 & 0.062 & 0.023 & -0.013 & 0.009 \\
AUX & 0.053 & 0.010 & 0.002 & 0.010 & 0.008 & 0.036 & 0.012 & 0.032 & 0.010 & 0.010 & 0.048 & 0.045 & 0.055 & 0.007 \\
CCONJ & 0.044 & 0.025 & 0.024 & 0.005 & 0.012 & \hlc[darkgreen!30]{0.043} & \hlc[darkgreen!50]{0.034} & -0.020 & 0.010 & 0.009 & \hlc[darkgreen!80]{0.165} & 0.042 & -0.007 & -0.023 \\
DET & 0.006 & 0.004 & 0.006 & 0.002 & -0.004 & 0.001 & 0.011 & \hlc[darkgreen!50]{0.043} & -0.007 & 0.002 & 0.046 & 0.018 & 0.011 & 0.008 \\
INTJ & -0.066 &  & -0.024 & 0.013 & 0.010 & -0.015 & -0.087 & 0.004 & \hlc[darkgreen!80]{0.037} & -0.019 & 0.031 & -0.041 & -0.009 &  \\
NOUN & 0.012 & -0.010 & 0.000 & 0.010 & -0.001 & 0.000 & -0.008 & 0.003 & -0.011 & -0.003 & 0.044 & -0.010 & -0.006 & -0.002 \\
NUM & 0.011 & -0.005 & -0.005 & -0.008 & 0.002 & 0.017 & \hlc[darkgreen!30]{0.019} & -0.046 & -0.002 & 0.009 & 0.008 & 0.025 & -0.000 & 0.004 \\
PART & 0.025 & -0.007 & 0.029 & 0.063 &  & -0.718 & 0.006 &  &  &  & 0.018 & 0.016 &  & -0.006 \\
PRON & 0.019 & 0.014 & -0.002 & 0.021 & 0.039 & 0.003 & -0.009 & \hlc[darkgreen!80]{0.047} & 0.006 & 0.013 & 0.029 & 0.023 & 0.000 & 0.023 \\
PRON.1 & 0.015 & 0.011 & 0.009 & 0.015 & 0.043 & 0.021 &  &  & 0.008 & 0.015 & 0.046 & 0.015 & -0.012 & 0.025 \\
PRON.1.Plur & 0.027 & 0.007 & -0.002 & 0.008 & \hlc[darkgreen!30]{0.082} & 0.004 &  &  &  & \hlc[darkgreen!30]{0.045} & 0.012 & 0.013 & -0.022 & 0.033 \\
PRON.1.Sing & -0.036 & 0.014 & 0.017 & 0.020 & 0.016 & 0.037 &  &  &  & 0.001 & \hlc[darkgreen!50]{0.075} & 0.015 & -0.006 &  \\
PRON.2 & 0.040 & \hlc[darkgreen!50]{0.222} & -0.020 & 0.037 & \hlc[darkgreen!50]{0.108} & 0.015 &  &  & 0.013 & \hlc[darkgreen!80]{0.171} & -0.017 & \hlc[darkgreen!50]{0.103} & -0.026 & 0.009 \\
PRON.2.Plur & \hlc[darkgreen!30]{0.075} & -0.055 & -0.019 & -0.008 & 0.088 & 0.011 &  &  &  &  & -0.008 & \hlc[darkgreen!30]{0.069} & -0.024 &  \\
PRON.2.Sing & 0.009 & \hlc[darkgreen!80]{0.226} & -0.021 & \hlc[darkgreen!80]{0.357} & \hlc[darkgreen!80]{0.125} & \hlc[darkgreen!50]{0.052} &  &  &  & & -0.033 & \hlc[darkgreen!80]{0.412} & -0.038 &  \\
PRON.3 & 0.018 & 0.026 & -0.009 & 0.024 & 0.031 & -0.020 &  &  & 0.004 & 0.033 & 0.029 & 0.042 & 0.008 & \hlc[darkgreen!50]{0.045} \\
PRON.3.Dual & 0.057 &  &  &  &  &  &  &  &  &  &  &  &  &  \\
PRON.3.Plur & 0.016 & 0.017 & -0.021 & 0.037 & 0.050 & 0.024 &  &  &  & \hlc[darkgreen!50]{0.058} & 0.062 & 0.038 & \hlc[darkgreen!80]{0.047} & \hlc[darkgreen!30]{0.038} \\
PRON.3.Sing & 0.017 & 0.032 & 0.000 & 0.030 & 0.026 & 0.009 &  &  &  & 0.014 & 0.046 & 0.044 & -0.001 &  \\
PRON.Plur &  & 0.001 & 0.018 & \hlc[darkgreen!30]{0.096} &  & 0.021 &  &  &  & 0.003 &  & -0.027 &  &  \\
PRON.Sing &  & 0.002 & -0.005 & 0.025 & -0.004 & 0.005 &  &  &  & 0.002 &  & 0.007 &  &  \\
PROPN & 0.016 & -0.014 & -0.002 & 0.018 & 0.017 & -0.016 & -0.018 & 0.003 & -0.005 & -0.013 & 0.007 & 0.021 & -0.014 & 0.005 \\
PUNCT & \hlc[darkgreen!50]{0.129} & 0.007 & 0.012 & 0.001 & 0.019 & 0.019 & \hlc[darkgreen!80]{0.353} & 0.017 & \hlc[darkgreen!50]{0.018} & 0.021 & 0.005 & 0.017 & \hlc[darkgreen!50]{0.022} & \hlc[darkgreen!80]{0.106} \\
SCONJ & \hlc[darkgreen!80]{0.137} & -0.001 & 0.017 & 0.001 & 0.007 & -0.000 & 0.004 & 0.005 & 0.005 & 0.003 & 0.044 & -0.001 &  &  \\
SYM & 0.050 & \hlc[darkgreen!30]{0.081} & \hlc[darkgreen!80]{0.136} & \hlc[darkgreen!50]{0.152} &  & 0.017 & -0.034 & -0.014 & -0.010 & -0.071 &  & -0.040 &  & 0.015 \\
VERB & 0.042 & 0.006 & 0.004 & 0.003 & 0.007 & 0.004 & 0.008 & \hlc[darkgreen!30]{0.036} & 0.002 & 0.005 & 0.047 & 0.015 & 0.014 & 0.015 \\
VERB.Fut &  &  & \hlc[darkgreen!50]{0.043} & 0.004 & 0.019 & 0.008 &  &  &  & -0.001 &  & -0.018 & 0.007 &  \\
VERB.Imp &  &  & \hlc[darkgreen!30]{0.039} & 0.010 &  & \hlc[darkgreen!80]{0.057} &  &  &  & 0.029 & \hlc[darkgreen!30]{0.069} &  &  &  \\
VERB.Past &  & 0.041 & 0.011 & 0.009 & 0.008 & 0.007 &  &  & -0.001 & 0.005 & -0.009 & 0.064 & 0.010 &  \\
VERB.Pres &  & 0.013 & 0.001 & -0.001 &  & -0.006 &  &  & \hlc[darkgreen!30]{0.011} & 0.014 & 0.039 & 0.002 & \hlc[darkgreen!30]{0.016} &  \\
%X & 0.042 & 0.024 &  & 0.131 &  & -0.013 &  & 0.028 & 0.179 & 0.242 &  &  &  & 0.019 \\
\midrule
ellipsis & 0.052 & -0.053 & -0.111 & 0.055 & 0.071 & 0.019 & 0.020 & 0.022 & 0.037 & -0.070 & 0.111 & -0.020 & -0.041 & 0.082 \\
formality &  & 0.038 & 0.077 & 0.040 &  & 0.048 & 0.036 & 0.022 & 0.014 & 0.008 & 0.008 & 0.107 & -0.073 & 0.012 \\
lexical & -0.006 & 0.003 & 0.011 & -0.001 & 0.003 & 0.001 & -0.007 & -0.008 & -0.004 & 0.002 & 0.034 & -0.002 & 0.008 & 0.004 \\
no tag & 0.041 & 0.001 & 0.003 & 0.005 & 0.005 & 0.006 & 0.011 & 0.013 & 0.002 & 0.005 & 0.036 & 0.009 & 0.003 & 0.017 \\
pronouns & 0.028 & 0.068 & -0.002 & 0.055 &  & 0.006 & -0.027 &  &  & 0.055 & 0.008 &  &  &  \\
verb form &  &  & 0.042 & 0.009 & 0.009 & 0.041 &  &  & -0.002 &  & 0.046 & 0.065 & 0.013 &  \\
with tag & -0.001 & 0.024 & 0.018 & 0.021 & 0.005 & 0.013 & 0.023 & 0.005 & 0.001 & 0.010 & 0.034 & 0.056 & 0.002 & 0.009 \\
\bottomrule
\end{tabular}}
\vspace{-1mm}
\caption{P-CXMI for all POS tags and our ambiguity tags. In the top two rows, CXMI is the average of P-CXMI for each sentence across the corpus, and P-CXMI is the average of P-CXMI over all tokens in the corpus. Per-tag values are the average of P-CXMI for each token with the tag. The 3 highest P-CXMI scores are highlighted in varying intensities of green.}
\vspace{-2mm}
\label{table:all_cxmi}
\end{table*}
Table \ref{table:all_cxmi} presents the average P-CXMI value per POS tag and per MuDA tag.

\section{Tag Numbers}
\label{appendix:total-counts}

\autoref{table:count_tags} lists the counts of each tag per language.

\begin{table}[t]
\centering
\begin{tabular}{c|ccccc}
\toprule
 & pronouns & formality & verb form  & lexical & ellipsis\\ 
 \midrule
ar & 90 & 0 & 0 & 116 & 982 \\
de & 398 & 1000 & 0 & 19 & 1356 \\
es & 245 & 86 & 409 & 15 & 1496 \\
fr & 1591 & 839 & 1938 & 48 & 1586 \\
he & 0 & 0 & 468 & 122 & 1210 \\
it & 182 & 118 & 484 & 31 & 1320 \\
ja & 245 & 3328 & 0 & 94 & 990 \\
ko & 0 & 221 & 0 & 71 & 373 \\
nl & 0 & 783 & 1060 & 27 & 1590 \\
pt\_br & 372 & 515 & 0 & 27 & 1677 \\
ro & 60 & 407 & 792 & 53 & 1002 \\
ru & 0 & 466 & 2091 & 41 & 668 \\
tr & 0 & 30 & 47 & 137 & 704 \\
zh\_cn & 0 & 526 & 0 & 49 & 1092 \\
\bottomrule
\end{tabular}
\vspace{-2mm}
\caption{Total number of MuDA tags on TED test data. '0' indicates that the phenomenon does not apply to that language.}
\vspace{-5mm}
\label{table:count_tags}
\end{table}

\section{Tagging other Document-level Datasets}
\label{appendix:other-datasets-tags}

\begin{figure}[!htbp]
    \centering
    \includegraphics[width=0.5\textwidth]{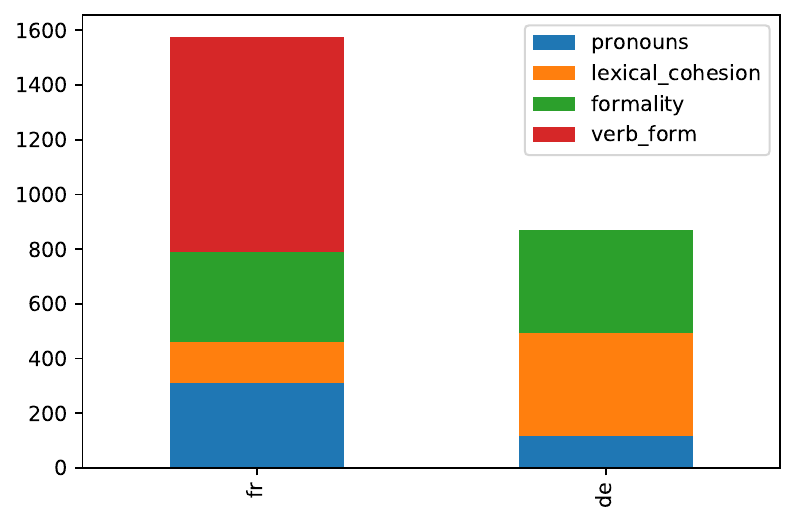}
    \caption{Number of tags for $\text{EN}\rightarrow \text{DE}$ and $\text{EN}\rightarrow \text{FR}$ in the IWSLT17 dataset. Lexical cohesion and verb form are common phenomena in this dataset.}
    \label{fig:iwslt-tags}
\end{figure}
\begin{figure}[!htbp]
    \centering
    \includegraphics[width=0.5\textwidth]{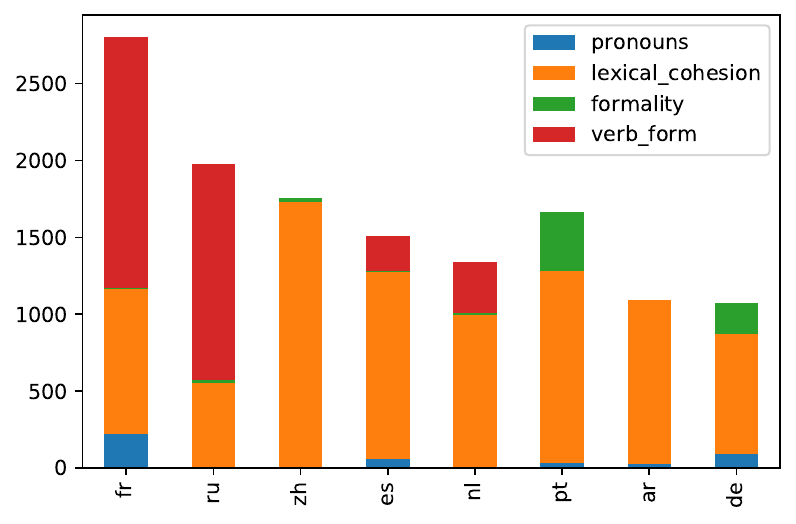}
    \caption{Number of tags across languages in the news-commentary dataset. Lexical cohesion and verb form are common phenomena in this dataset.}
    \label{fig:news-tags}
\end{figure}

We report the number of tags found for two other document-level datasets commonly used in the literature: (1) IWSLT-17 \cite{cettolo-2012} test sets for  $\text{EN}\rightarrow \text{DE}$ and $\text{EN}\rightarrow \text{FR}$ and (2) A randomly subsampled portion of the news-commentary dataset for $\text{EN} \rightarrow \{ \text{AR}, \text{DE}, \text{ES}, \text{FR}, \text{NL}, \text{PT}, \text{RU}, \text{ZH}\}$ \cite{news-commentary}. These results can be found respectively in \autoref{fig:iwslt-tags} and \autoref{fig:news-tags}.

\section{Tagger Details}
\label{appendix:tagger}

\subsection{Formality Words}
Table \ref{table:formality} gives the list of words related to formality for each target language.

\begin{table*}[t]
\resizebox{\linewidth}{!}{ 
\begin{tabular}{c|c}
\toprule
\multirow{2}{*}{de} & du\\
& sie \\
\midrule
\multirow{2}{*}{es}& tú, tu, tus, ti, contigo, tuyo, te, tuya \\
& usted, vosotros, vuestro, vuestra, vuestras, os \\
\midrule
\multirow{2}{*}{fr}&	tu, ton,ta, tes, toi, te, tien, tiens, tienne, tiennes \\
& vous, votre, vos \\
\midrule
\multirow{2}{*}{it}&tu, tuo, tua, tuoi \\
& lei, suo, sua, suoi \\
\midrule
\multirow{2}{*}{ja}& \begin{CJK}{UTF8}{min} だ, だっ, じゃ, だろう, だ, だけど, だっ \end{CJK}\\
& \begin{CJK}{UTF8}{min}ござい, ます, いらっしゃれ, いらっしゃい, ご覧, 伺い, 伺っ, 存知, です, まし\end{CJK} \\
\midrule
\multirow{2}{*}{ko}&\begin{CJK}{UTF8}{mj} 제가, 저희, 나\end{CJK} \\
& \begin{CJK}{UTF8}{mj} 댁에, 성함, 분, 생신, 식사, 연세, 병환, 약주, 자제분, 뵙다, 저 \end{CJK} \\
\midrule
\multirow{2}{*}{nl} & jij, jouw, jou, jullie, je \\
& u, men, uw \\
\midrule
\multirow{2}{*}{pt}& tu, tua, teu, teus, tuas, te \\
& você, sua, seu, seus, suas, lhe \\
\midrule
\multirow{2}{*}{ro}&tu, el, ea, voi,  ei, ele, tău, ta, tale, tine\\
& dumneavoastră, dumneata, mata,matale,dânsul, dânsa dumnealui,dumneaei, dumnealor \\
\midrule
\multirow{2}{*}{ru} & \foreignlanguage{russian}{ты, тебя, тебе, тобой, твой, твоя, твои,тебе} \\
& \foreignlanguage{russian}{вы, вас, вам, вами, ваш, ваши}\\
\midrule
\multirow{2}{*}{tr}&sen, senin\\
& siz, sizin \\
\midrule
\multirow{2}{*}{zh} &\begin{CJK*}{UTF8}{gbsn} 你 \end{CJK*} \\ 
&\begin{CJK*}{UTF8}{gbsn} 您 \end{CJK*} \\
\bottomrule
\end{tabular}}
\vspace{-1mm}
\caption{Words related to formality for each target language.}
\vspace{-2mm}
\label{table:formality}
\end{table*}

\subsection{Ambiguous Pronouns}
\label{appendix:pronouns}

Table \ref{table:pronouns} provides English pronouns and the list of possible target pronouns.

\begin{table}[t]
\centering
\resizebox{0.7\linewidth}{!}{ 
\begin{tabular}{cc|c}
\toprule
\multirow{3}{*}{ar} & you & \RL{انت, انتَ, انتِ, انتى, أنتم, أنتن, انتو, أنتما, أنتما} \\
& it& \RL{هو, هي} \\
& they, them& \RL{هم, هن, هما} \\
\midrule
\multirow{1}{*}{de} & it & er, sie, es\\
\midrule
\multirow{6}{*}{es}& it & él, ella \\
& they, them & ellos, ellas \\
& this & ésta, éste, esto\\
& that & esa, ese \\
& these & estos, estas \\
& those & aquellos, aquellas, ésos, ésas\\
\midrule
\multirow{7}{*}{fr}& it & il, elle, lui \\
& they, them & ils, elles \\
& we & nous, on \\
& this & celle, ceci\\
& that & celle, celui \\
& these, those & celles, ceux \\
\midrule
\multirow{6}{*}{it}&it & esso, essa \\
& this & questa, questo\\
& that & quella, quello \\
& these & queste, questi \\
& those & quelle, quelli\\
\midrule
\multirow{1}{*}{ja}& I & \begin{CJK}{UTF8}{min} 私, 僕, 俺\end{CJK}\\
\midrule
\multirow{5}{*}{pt}& it & ele, ela, o, a\\
& them & eles, elas, os, as \\
& they & eles, elas \\
& this, that & este, esta, esse, essa \\
& these, those & estes, estas, esses, essas\\
\midrule
\multirow{2}{*}{ro} & it & el, ea\\
& they, them & ei, ele \\
\bottomrule
\end{tabular}}
\vspace{-1mm}
\caption{Ambiguous pronouns w.r.t. English for each target language.}
\vspace{-2mm}
\label{table:pronouns}
\end{table}

\subsection{Ambiguous Verbs}
\label{appendix:verbs}

Table \ref{table:verbs} lists verb forms that may require disambiguation during translation.

\begin{table}[t]
\centering
\resizebox{0.5\linewidth}{!}{ 
\begin{tabular}{c|c}
\toprule
{es}& Imperfect, Pluperfect, Future \\
\midrule
{fr}& Imperfect, Past, Pluperfect \\
\midrule
he & Imperfect, Future, Pluperfect \\
\midrule
it&Imperfect, Pluperfect, Future \\
\midrule
{nl} & Past \\
\midrule
\multirow{1}{*}{pt}& Pluperfect \\
\midrule
\multirow{1}{*}{ro}&Imperfect, Past, Future \\
\midrule
\multirow{1}{*}{ru} & Past\\
\midrule
\multirow{1}{*}{tr}&Pluperfect\\
\bottomrule
\end{tabular}}
\vspace{-1mm}
\caption{Ambiguous verb forms w.r.t. English for each target language.}
\vspace{-2mm}
\label{table:verbs}
\end{table}

\subsection{Ellipsis Classifier}
\label{appendix:ellipsis}
We train a BERT text classification model \cite{devlin2018bert} on data from the Penn Treebank, where we labeled each sentence containing the tag `*?*' as containing ellipsis \cite{bies1995bracketing}. We obtain 248,596 sentences total, with 2,863 tagged as ellipsis. Then, our model using HuggingFace Transformers \cite{wolf-etal-2020-transformers}. To address the imbalance in labels, we up-weight the loss for samples tagged as ellipsis by a factor of 100.

\section{Training details}
\label{appendix:training_details}

The \textit{transformer-small} model has hidden size of 512, feedforward size of 1024, 6 layersa and 8 attention heads. 
The \textit{transformer-large} model has hidden size of 1024, feedforward size of 4096, 6 layers, 16 attention heads.

As in \citet{46201}, we train using the Adam optimizer with $\beta_1=0.9$ and $\beta_2=0.98$ and use an inverse square root learning rate scheduler, with an initial value of $10^{-4}$ for \texttt{large} model and $5\times 10^{-4}$ for the \texttt{base} and \texttt{multi} models, with a linear warm-up in the first $4000$ steps. 

For the pretrained models we used Paracrawl \cite{espla-etal-2019-paracrawl} for German and French, JParacrawl \cite{morishita-etal-2020-jparacrawl} for Japanese and the Backtranslated News from WMT2021 for Chinese. 

Due to the sheer number of experiments, we use a single seed per experiment.

We base our experiments on the framework \textit{Fairseq} \cite{ott-etal-2019-fairseq}. 

\section{Results Tables}
\label{appendix:results_tables}

\begin{table*}[t]
\centering
\resizebox{0.95\linewidth}{!}{ 
\begin{tabular}{cc|cccccccccccccc}
\toprule
 && ar&de&es&fr&he&it&ja&ko&nl&pt&ro&ru&tr&zh\\
 \midrule
\multirow{3}{*}{BLEU} &no-context & 17.25 &28.02 &35.72 &37.74 &32.70 &32.30 & 7.10 &6.80 &32.22 &39.03 &25.36 &17.00 &12.32 &15.96 \\
&context &16.92 &28.24 &36.00 &37.23 &32.92 &32.11 &4.48 &3.77 &32.67 &39.10 &25.37 &17.14 &11.97 &15.01 \\
&context-gold & \underline{18.61} & \underline{28.60} & \underline{36.27} & 37.96 & \underline{33.41} & 32.37 &5.96 & 6.92 & \underline{32.73} & \underline{39.55} & \underline{28.49} & \underline{17.70} & 12.49 & 16.05 \\
\midrule
\multirow{3}{*}{COMET} &no-context &0.0002 &0.1841 &0.3809 & 0.3087 & 0.0948 & 0.2608 & -0.5366 & -0.0275 &0.3105 & 0.4562 & 0.3826 & 0.0033 & 0.2113 &-0.1419 \\
&context & -0.0066 &0.1846 &0.3875 &0.2811 &0.0887 &0.2496 &-0.7728 &-0.3339 & 0.3238 &0.4444 &0.3747 &-0.0190 &0.1831 &-0.1917 \\
&context-gold & 0.0025 & 0.1886 & 0.3879 &0.2821 &0.0922 &0.2467 &-0.6827 &-0.1000 &0.3218 &0.4506 &0.3805 &-0.0173 &0.1871 & -0.1274 \\
\midrule
% \midrule
% \multirow{3}{*}{all} & no-context & 0.477&0.612&0.665&0.663&0.598&0.621&0.338&0.301&0.646&0.686&0.552&0.469&0.432&0.372 \\
% & context & 0.474&0.615&0.665&0.661&0.600&0.623&0.267&0.200&0.648&0.686&0.554&0.472&0.432&0.333 \\
% & context-gold & \underline{0.495}&\underline{0.618}&0.667&0.666&\underline{0.603}&\underline{0.625}&0.313&0.289&0.649&0.689&\underline{0.586}&\underline{0.477}&0.433&\underline{0.382} \\
\midrule
\multirow{3}{*}{ellipsis} & no-context & 0.374&0.387&0.210&0.400&0.439&0.259&0.123&0.169&0.400&0.342&0.333&0.255&0.165&0.145 \\
& context & 0.325&0.323&0.333&0.406&0.389&0.400&0.021&0.033&0.471&0.450&0.270&0.292&0.240&0.135 \\
& context-gold & 0.388&0.296&0.300&0.435&0.371&0.381&0.025&0.150&0.444&0.450&0.306&0.226&0.187&0.154 \\
\midrule
\multirow{3}{*}{formality} & no-context & \textendash&0.607&0.370&0.792&\textendash&0.429&0.443&0.399&0.682&0.599&0.434&0.464&0.097&0.691 \\
& context & \textendash&\underline{0.639}&0.351&0.791&\textendash&0.462&0.414&0.397&0.694&0.600&0.405&0.469&0.083&0.695 \\
& context-gold & \textendash&\underline{0.661}&0.443&0.803&\textendash&0.464&0.431&0.425&0.697&0.622&0.440&0.492&0.182&\underline{0.741} \\
\midrule
\multirow{3}{*}{lexical} & no-context & 0.639&0.762&0.819&0.826&0.723&0.766&0.615&0.574&0.821&0.853&0.661&0.624&0.671&0.645 \\
& context & 0.630&0.736&0.833&0.830&0.722&0.772&0.572&0.524&0.825&0.851&\underline{0.689}&0.624&0.647&0.644 \\
& context-gold & \underline{0.675}&0.737&0.832&0.832&0.727&0.773&0.614&0.593&0.828&0.857&\underline{0.713}&0.625&0.647&\underline{0.676} \\
\midrule
\multirow{3}{*}{pronouns} & no-context & 0.660&0.613&0.576&0.774&\textendash&0.548&0.473&\textendash&\textendash&0.452&0.356&\textendash&\textendash&\textendash \\
& context & 0.691&0.614&0.538&0.771&\textendash&0.549&0.377&\textendash&\textendash&0.451&0.414&\textendash&\textendash&\textendash \\
& context-gold & 0.700&0.624&0.550&0.788&\textendash&0.530&0.428&\textendash&\textendash&0.485&0.432&\textendash&\textendash&\textendash \\
\midrule
\multirow{3}{*}{verb tense} & no-context & \textendash&\textendash&0.263&0.435&0.227&0.308&\textendash&\textendash&0.477&\textendash&0.292&0.215&0.128&\textendash \\
& context & \textendash&\textendash&0.287&0.442&0.229&0.282&\textendash&\textendash&0.479&\textendash&0.292&0.215&0.094&\textendash \\
& context-gold & \textendash&\textendash&0.272&0.435&0.229&0.285&\textendash&\textendash&0.487&\textendash&0.328&\underline{0.238}&0.120&\textendash \\
\midrule
\bottomrule
\end{tabular}}
\vspace{-2mm}
\caption{BLEU, COMET, and Word f-measure per tag for \texttt{base} context-aware models. BLEU, COMET and word f-measures statistically significantly higher than no-context ($p$ < 0.05) are \ul{underlined}.}
\vspace{-5mm}
\label{table:fmeas}
\end{table*}

\begin{table}[t!]
\centering
\resizebox{.5\linewidth}{!}{ 
\begin{tabular}{cc|cccccccccccccc}
\toprule
& &de&fr&ja&zh\\
\midrule
\multirow{3}{*}{BLEU} &no-context & 36.09 &45.64 &15.55 &22.15 \\
&context &35.86 &45.40 &12.68 &\underline{22.68} \\
&context-gold &  \underline{36.69} & \underline{46.60} & \underline{16.60} & \underline{22.98} \\
\midrule
\multirow{3}{*}{COMET} &no-context &0.5256 &0.6332 &0.0602 &0.1160 \\
&context &0.5337 &0.6425 &0.0753 &\underline{0.2705} \\
&context-gold & \underline{0.5427} & \underline{0.6529} & \underline{0.1808} & \underline{0.2809} \\
\midrule
\midrule
% \multirow{3}{*}{all} & no-context & 0.669&0.714&0.456&0.419 \\
% & context & 0.667&0.713&0.401&\underline{0.431} \\
% & context-gold & \underline{0.675}&\underline{0.720}&0.458&\underline{0.442} \\
% \midrule
\multirow{3}{*}{ellipsis} & no-context & 0.429&0.462&0.126&0.254 \\
& context & 0.518&0.393&0.068&0.230 \\
& context-gold & 0.444&0.444&0.144&0.209 \\
\midrule
\multirow{3}{*}{formality} & no-context & 0.642&0.824&0.510&0.747 \\
& context & 0.640&0.810&0.513&0.739 \\
& context-gold & \underline{0.692}&0.820&\underline{0.537}&0.739 \\
\midrule
\multirow{3}{*}{lexical} & no-context & 0.773&0.864&0.704&0.661 \\
& context & 0.776&0.868&0.699&0.671 \\
& context-gold & \underline{0.796}&\underline{0.875}&\underline{0.740}&\underline{0.696} \\
\midrule
\multirow{3}{*}{pronouns} & no-context & 0.633&0.790&0.493&\textendash \\
& context & 0.635&0.795&0.541&\textendash \\
& context-gold & 0.665&0.801&0.536&\textendash \\
\midrule
\multirow{3}{*}{verb tense} & no-context & \textendash&0.526&\textendash&\textendash \\
& context & \textendash&0.532&\textendash&\textendash \\
& context-gold & \textendash&0.534&\textendash&\textendash \\
\midrule
\bottomrule
\end{tabular}}
\vspace{-1mm}
\caption{Word f-measure per tag for \texttt{large} models. BLEU, COMET, word f-measures statistically significantly higher than no-context ($p$ < 0.05) are \ul{underlined}.} 
\vspace{-6mm}
\label{table:pre-fmeas}
\end{table}

\begin{table*}[htbp!]
\centering
\resizebox{0.97\linewidth}{!}{ 
\begin{tabular}{cc|cccccccccccccc}
\toprule
 && ar&de&es&fr&he&it&ja&ko&nl&pt&ro&ru&tr&zh\\
\midrule
\multirow{3}{*}{BLEU} &Google &11.73 &34.76 &43.47 &30.77 &10.77 &31.34 &12.98 &8.77 &38.51 &38.49 &28.54 &24.79 &18.22 &28.92 \\
&DeepL (sent) & x &34.29 &42.00 &42.57 & x &35.41 &14.88 & x &37.58 &37.37 &28.98 &25.67 & x &27.94 \\
&DeepL (doc) & x & \underline{36.75} & \underline{43.06} &\underline{43.43} & x &\underline{36.04} &\underline{15.66} & x &\underline{38.29} &\underline{37.76} &\underline{29.79} &\underline{26.53} & x &27.34 \\
\midrule
\multirow{3}{*}{COMET} &Google &0.3862 &0.5480 &0.7694 &0.6655 &0.3666 &0.6707 &0.2116 &0.4721 &0.6401 &0.7925 &0.7437 &0.5121 &0.7254 &0.3697 \\
&DeepL (sent) & x &0.5750 &0.7680 &0.7121 & x &0.6951 &0.2973 & x &0.6321 &0.7513 &0.8026 &0.5501 & x &0.3739 \\
&DeepL (doc) & x &\underline{0.5848} &\underline{0.7882} &\underline{0.7267} & x &\underline{0.7049} &0.2343 & x &0.6357 &0.7572 &0.8121 &0.5495 & x &0.2453 \\
\midrule
\midrule
% \multirow{3}{*}{all} & Google & 0.512&0.667&0.719&0.662&0.594&0.663&0.444&0.369&0.692&0.723&0.591&0.547&0.514&0.513 \\
% & DeepL (sent) & x&0.662&0.709&0.709&x&0.688&0.456&x&0.681&0.709&0.596&0.554&x&0.507 \\
% & DeepL (doc) & x&\underline{0.680}&\underline{0.719}&\underline{0.716}&x&\underline{0.694}&0.459&x&\underline{0.687}&\underline{0.716}&\underline{0.605}&\underline{0.562}&x&0.479 \\
% \midrule
\multirow{3}{*}{ellipsis} & Google & 0.343&0.667&0.500&0.306&0.359&0.468&0.279&0.352&0.389&0.632&0.405&0.367&0.236&0.323 \\
& DeepL (sent) & x&0.417&0.400&0.422&x&0.500&0.275&x&0.500&0.421&0.458&0.385&x&0.303 \\
& DeepL (doc) & x&0.435&0.526&0.493&x&0.553&0.208&x&0.500&0.359&0.532&0.385&x&0.295 \\
\midrule
\multirow{3}{*}{formality} & Google & \textendash&0.621&0.404&0.738&\textendash&0.458&0.489&0.300&0.638&0.633&0.479&0.512&0.367&0.599 \\
& DeepL (sent) & \textendash&0.641&0.419&0.733&\textendash&0.455&0.487&x&0.610&0.625&0.533&0.533&x&0.729 \\
& DeepL (doc) & \textendash&0.670&0.446&\underline{0.785}&\textendash&0.503&\underline{0.520}&x&0.641&0.614&0.526&0.534&x&0.664 \\
\midrule
\multirow{3}{*}{lexical} & Google & 0.665&0.786&0.854&0.827&0.697&0.794&0.602&0.611&0.825&0.860&0.700&0.635&0.677&0.693 \\
& DeepL (sent) & x&0.773&0.840&0.860&x&0.805&0.657&x&0.799&0.848&0.714&0.653&x&0.660 \\
& DeepL (doc) & x&0.776&0.841&\underline{0.872}&x&0.812&0.640&x&0.802&0.846&0.713&0.649&x&0.657 \\
\midrule
\multirow{3}{*}{pronouns} & Google & 0.670&0.648&0.626&0.757&\textendash&0.511&0.486&\textendash&\textendash&0.488&0.326&\textendash&\textendash&\textendash \\
& DeepL (sent) & x&0.608&0.538&0.737&\textendash&0.543&0.526&\textendash&\textendash&0.483&0.394&\textendash&\textendash&\textendash \\
& DeepL (doc) & x&\underline{0.706}&\underline{0.588}&\underline{0.789}&\textendash&0.551&0.557&\textendash&\textendash&0.513&0.472&\textendash&\textendash&\textendash \\
\midrule
\multirow{3}{*}{verb tense} & Google & \textendash&\textendash&0.415&0.529&0.311&0.450&\textendash&\textendash&0.554&\textendash&0.358&0.314&0.167&\textendash \\
& DeepL (sent) & \textendash&\textendash&0.390&0.553&x&0.478&\textendash&\textendash&0.562&\textendash&0.400&0.327&x&\textendash \\
& DeepL (doc) & \textendash&\textendash&\underline{0.426}&0.562&x&0.445&\textendash&\textendash&0.567&\textendash&0.411&\underline{0.349}&x&\textendash \\
\midrule
\bottomrule
\end{tabular}}
\vspace{-1.5mm}
\caption{Scores for commercial models. DeepL (doc) BLEU, COMET and word f-measures statistically significantly higher than DeepL (sent) are \ul{underlined}.} 
\vspace{-5mm}
\label{table:prov-fmeas}
\end{table*}

% \section{Test Appendix}
% \label{appendix:training_details}

\end{document}